\pdfoutput=1
\documentclass[conference]{IEEEtran}
\IEEEoverridecommandlockouts

\usepackage{cite}
\usepackage{amsmath,amssymb,amsfonts}
\usepackage{algorithmic}
\usepackage{graphicx}
\usepackage{textcomp}
\usepackage{xcolor}
\usepackage{subfig}
\usepackage{soul}

\def\BibTeX{{\rm B\kern-.05em{\sc i\kern-.025em b}\kern-.08em
    T\kern-.1667em\lower.7ex\hbox{E}\kern-.125emX}}

\begin{document}

\title{Multi-UAV Active Sensing with Information Gain-based Planning and Belief Fusion}

\author{
\IEEEauthorblockN{Sara Habibi, Lino Marques}
\IEEEauthorblockA{
\textit{Institute of Systems and Robotics} \\
University of Coimbra, 3030-290 Coimbra, Portugal \\
{sara.habibi, lino}@isr.uc.pt
}
}

\maketitle

\begin{abstract}
Unmanned aerial vehicles (UAVs) are increasingly used for active sensing and information gathering in spatially distributed environments. Their performance, however, is constrained by limited flight time, sensing uncertainty, and the trade-off between spatial coverage and observation accuracy. This paper presents a real-world validation of a multi-UAV active sensing framework for probabilistic binary terrain mapping, with precision agriculture used as the application case. The environment is represented as a probabilistic belief map, where spatial dependencies are modeled through a factor-graph formulation. UAV decision making is guided by Information Gain based Informative Path Planning (IGbIPP), and the approach is compared with Random Walk and Sweep coverage path planning baselines using both synthetic terrains and real UAV-derived agricultural imagery. The study also evaluates spatial correlation weights and several probabilistic belief-fusion rules for multi-UAV information sharing. Results show that IGbIPP reduces entropy and mapping error more effectively than the baselines, while a wider field of view improves real-world coverage and map accuracy. The results further show that simple equal or biased spatial weights can be more robust than adaptive weights, and that Bayesian, log-odds, and Dempster--Shafer fusion achieve the best cooperative mapping performance. These findings highlight the importance of uncertainty-driven planning, sensing geometry, spatial modeling, and probabilistic fusion for real-world UAV-based active sensing.
\end{abstract}


\section{Introduction}

Unmanned aerial vehicles (UAVs) have become an important platform for active sensing and information gathering in spatially distributed environments. They are increasingly used in applications such as precision agriculture, environmental monitoring, infrastructure inspection, disaster response, healthcare support, and search-and-rescue \cite{b1,b2,b19}. In these applications, the objective is not only to collect images, but also to decide where future observations should be acquired \cite{b5,b7}. Compared with fixed sensors, UAVs provide mobility, flexible viewpoints, and rapid deployment. However, their sensing performance is constrained by limited flight time, battery capacity, communication range, sensor footprint, and observation uncertainty. Therefore, efficient UAV sensing requires decision-making strategies that balance exploration, sensing accuracy, and spatial coverage.

A central challenge in UAV active sensing is the trade-off between coverage and observation quality. Flying at a higher altitude increases the camera footprint and allows a larger region to be observed at each step, but it may reduce spatial resolution and increase classification uncertainty. In contrast, flying at a lower altitude improves observation quality but reduces the covered area and increases the time required to monitor the full environment. Classical coverage path planning methods, such as sweep or boustrophedon-style trajectories, provide systematic coverage of a target area, but they usually do not adapt their motion to the current uncertainty of the estimated map \cite{b10}. Active sensing and informative path planning address this limitation by selecting future viewpoints according to information-theoretic objectives, such as entropy reduction or expected information gain.

This paper studies UAV active sensing for probabilistic terrain mapping, with precision agriculture used as the application case. In particular, the monitored field is represented as a binary vegetation/no-vegetation terrain, and the UAV maintains a probabilistic belief map over the environment. The objective is to reduce map uncertainty and mapping error by choosing informative sensing actions. Although the experimental dataset is agricultural, the proposed formulation is general and can be applied to other binary or semantic terrain-monitoring tasks where UAVs must reason under sensing uncertainty.

The study evaluates both single-UAV and multi-UAV active sensing. The single-UAV case compares Information Gain (IG) planning with Random Walk and Sweep baselines using synthetic terrains and real UAV-derived agricultural imagery. The multi-UAV case investigates how information sharing and probabilistic belief fusion affect cooperative terrain mapping. The real-world validation is performed using UAV imagery from the WeedMap dataset, which is converted into a binary vegetation/no-vegetation terrain map. This allows the framework to be tested not only on controlled synthetic terrains, but also on agricultural imagery containing real spatial structure and image variability.

The main contributions of this paper are summarized as follows:

\begin{itemize}
    \item A UAV active sensing framework for probabilistic binary terrain mapping, evaluated using both synthetic terrains and real UAV-derived agricultural imagery;

    \item A comparison of Random Walk, Sweep coverage path planning, and Information Gain planning under altitude-dependent sensing uncertainty;

    \item An analysis of spatial correlation modeling through equal, biased, and adaptive pairwise factor weights for belief-map refinement;

    \item An evaluation of multi-UAV information sharing using different probabilistic belief-fusion rules, including Bayesian fusion, weighted average, log-odds fusion, confidence-weighted fusion, Dempster--Shafer-style fusion, covariance-intersection-inspired fusion, and consensus fusion.
\end{itemize}

The remainder of this paper is organized as follows. Section II reviews related work on UAV active sensing, informative path planning, coverage planning, and agricultural monitoring. Section III presents the methodology. Section IV describes the experimental setup and results. Finally, Section V concludes the paper and discusses future work.

\section{Related Work}

UAV-based monitoring has been widely studied in precision agriculture and remote sensing. Low-altitude UAV remote sensing provides high spatial and temporal resolution, making it suitable for crop monitoring, weed detection, vegetation mapping, and field inspection \cite{b1,b2}. Multi-robot and multi-UAV systems further extend these capabilities by enabling faster coverage, distributed sensing, and cooperative information gathering \cite{b3}. However, agricultural UAV monitoring remains challenging because sensing quality depends on altitude, camera footprint, illumination, vegetation structure, and mission duration. These challenges motivate planning methods that can actively select informative viewpoints rather than only following precomputed coverage paths.

Active sensing and informative path planning provide a decision-making framework for this type of problem. Instead of treating all regions as equally important, informative planning selects actions that are expected to reduce uncertainty or improve the estimated map \cite{b5,b6}. Popovi\'c et al. developed UAV-based informative path planning methods for active classification and terrain monitoring, showing that information-theoretic planning can focus sensing effort on uncertain or valuable regions and can outperform traditional lawnmower-style coverage in mapping tasks \cite{b7,b8}. Hitz et al. also studied adaptive continuous-space informative path planning for environmental monitoring, demonstrating the value of optimizing sensing trajectories according to expected information gain \cite{b16}. These works provide the general basis for information-driven UAV sensing, but many evaluations are still performed primarily in simulation or under simplified environmental assumptions.

Coverage path planning remains an important baseline for UAV sensing because it provides predictable and systematic area coverage. General coverage path planning has been extensively reviewed in robotics \cite{b17}, and agricultural implementations have been developed for field operations and autonomous agricultural vehicles \cite{b18}. Boustrophedon and sweep-based patterns are especially common because they are simple, repeatable, and effective when full coverage is required. However, coverage planners usually do not explicitly reason about map uncertainty or spatially varying information value. As a result, they may spend sensing effort in already-known or low-value regions, while active sensing methods can adapt their motion based on the current belief map.

In precision agriculture, UAV imagery has been used for semantic crop and weed mapping. The WeedMap dataset provides large-scale multispectral UAV imagery for crop and weed segmentation in sugar beet fields and is widely used as a benchmark for agricultural mapping \cite{b11}. Previous studies have also applied active informative planning to UAV-based weed mapping, showing that map representation and uncertainty modeling strongly affect sensing efficiency \cite{b13}. Other works have combined semantic segmentation and coverage planning to improve agricultural drone missions in weed-infested fields \cite{b12,b14}. These studies show the importance of UAV-based agricultural perception, but the connection between real UAV-derived terrain maps, probabilistic belief updates, active sensing, and multi-UAV belief fusion remains less explored.

Multi-UAV active sensing introduces additional challenges related to communication, redundancy, and fusion of uncertain information. When several UAVs observe the same environment, they must coordinate their sensing actions and combine local beliefs without over-counting repeated information. Pierdicca et al. proposed a multi-UAV active sensing framework for precision agriculture using information gain, Bayesian fusion, and a factor-graph representation of the field \cite{b4}. Westheider et al. investigated deep reinforcement learning for multi-UAV adaptive path planning under team-size and communication constraints \cite{b9}. These studies demonstrate the potential of cooperative sensing, but further evaluation is needed to understand how different belief-sharing and fusion rules behave on real UAV-derived agricultural maps.

Building on this literature, the present work studies UAV active sensing as a general probabilistic terrain-mapping problem, while using agricultural vegetation mapping as the real-world application. The proposed evaluation compares information-gain planning with both random and systematic coverage baselines, analyzes the role of spatial correlation weights, and studies multi-UAV information sharing using several probabilistic belief-fusion rules.

\section{Methodology}

This work addresses autonomous agricultural field monitoring using single and multiple UAVs. The objective is to efficiently reduce map uncertainty and identify binary terrain features, such as vegetation and no-vegetation regions. The framework combines probabilistic terrain mapping, altitude-dependent sensing, information-gain-based planning, and multi-UAV information sharing.

\subsection{Environment and Belief Map}

We assume that the monitored field is represented as a grid-based binary terrain composed of equally sized square cells. Each cell \(m_i \in \{0,1\}\) indicates the absence or presence of a target feature. Since the true terrain state is not directly known, the UAVs maintain a probabilistic belief map \(B\), where each value \(B_i\) represents the probability that cell \(m_i\) contains the target feature.

The UAV configuration space is discretized into a three-dimensional lattice, allowing each UAV to move horizontally and vertically during the mission. At each time step, a UAV observes a subset of terrain cells inside its sensing footprint and updates the corresponding belief values using noisy sensor measurements. The updated belief map is then used by the planner to select future sensing actions.

Two types of terrain are considered in this work. First, simulated binary terrains \cite{b4} are used to reproduce and analyze the information-gain-based active sensing framework under controlled spatial correlation conditions. Second, real UAV agricultural imagery \cite{b11} is converted into binary vegetation or no-vegetation terrain maps to evaluate the behavior of the framework under realistic image variability.

\subsection{Sensor Model}

Given a UAV position and altitude, the sensor observes a patch of the terrain referred to as the sensing footprint. Low-altitude observations provide higher sensing accuracy but cover a smaller area, while high-altitude observations increase coverage at the cost of higher uncertainty. We denote the footprint bounding box associated with a UAV position \(x\) as \(fp(x)\), and the subset of terrain cells covered by the footprint as \(C(x)\). Assuming a bounding box footprint, its side length \(d\) depends on the UAV altitude \(h\) and the camera field of view (FoV):

\begin{equation}
d(h) = 2h \tan \left(\frac{\mathrm{FoV}}{2}\right)
\label{eq:footprint}
\end{equation}

\begin{figure}[t]
    \centering
    \subfloat[] {
        \includegraphics[width=0.45\columnwidth]{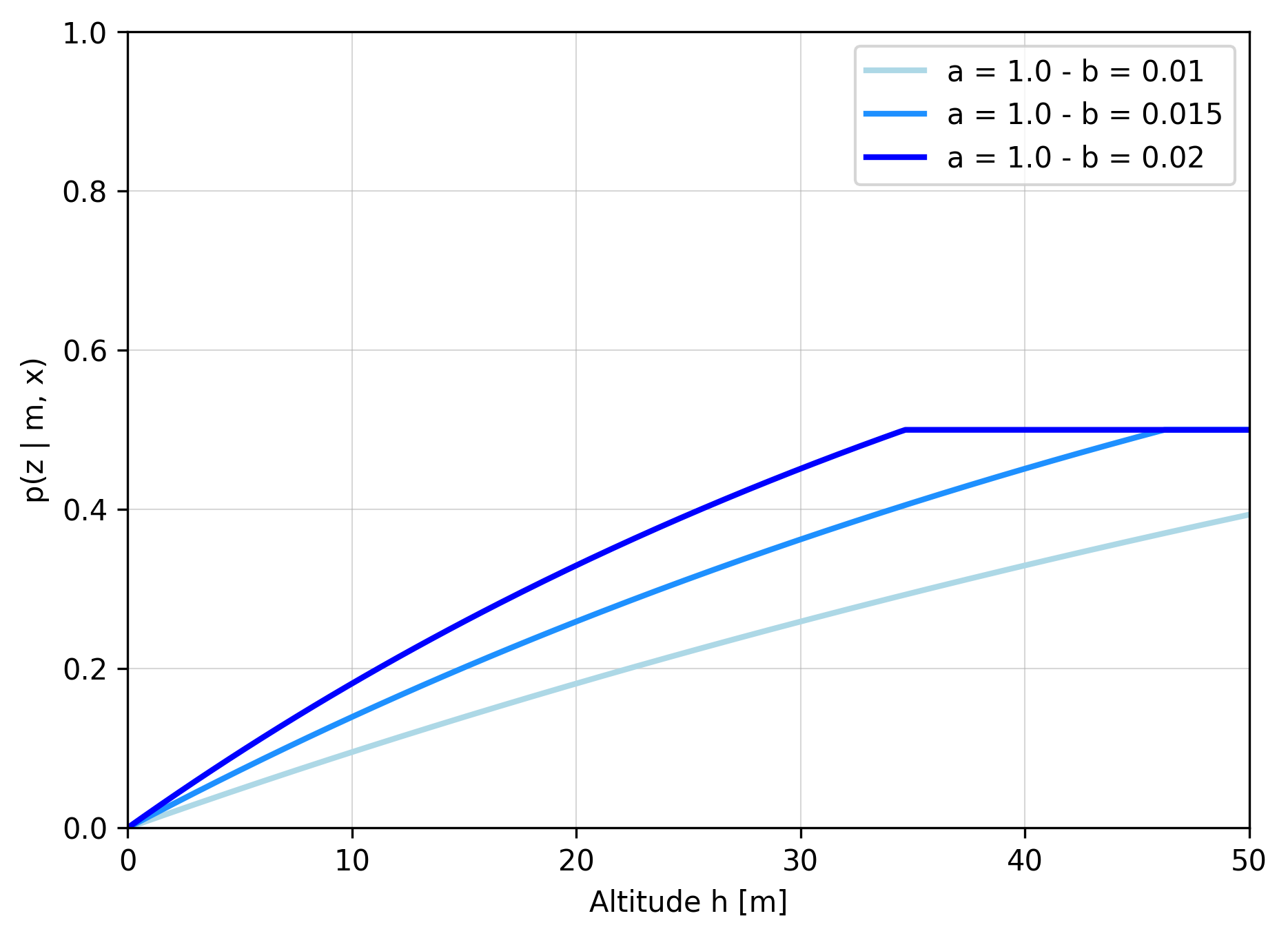}
        \label{fig:sensor_model}
    }
    \hfill
    \subfloat[] {
        \includegraphics[width=0.45\columnwidth]{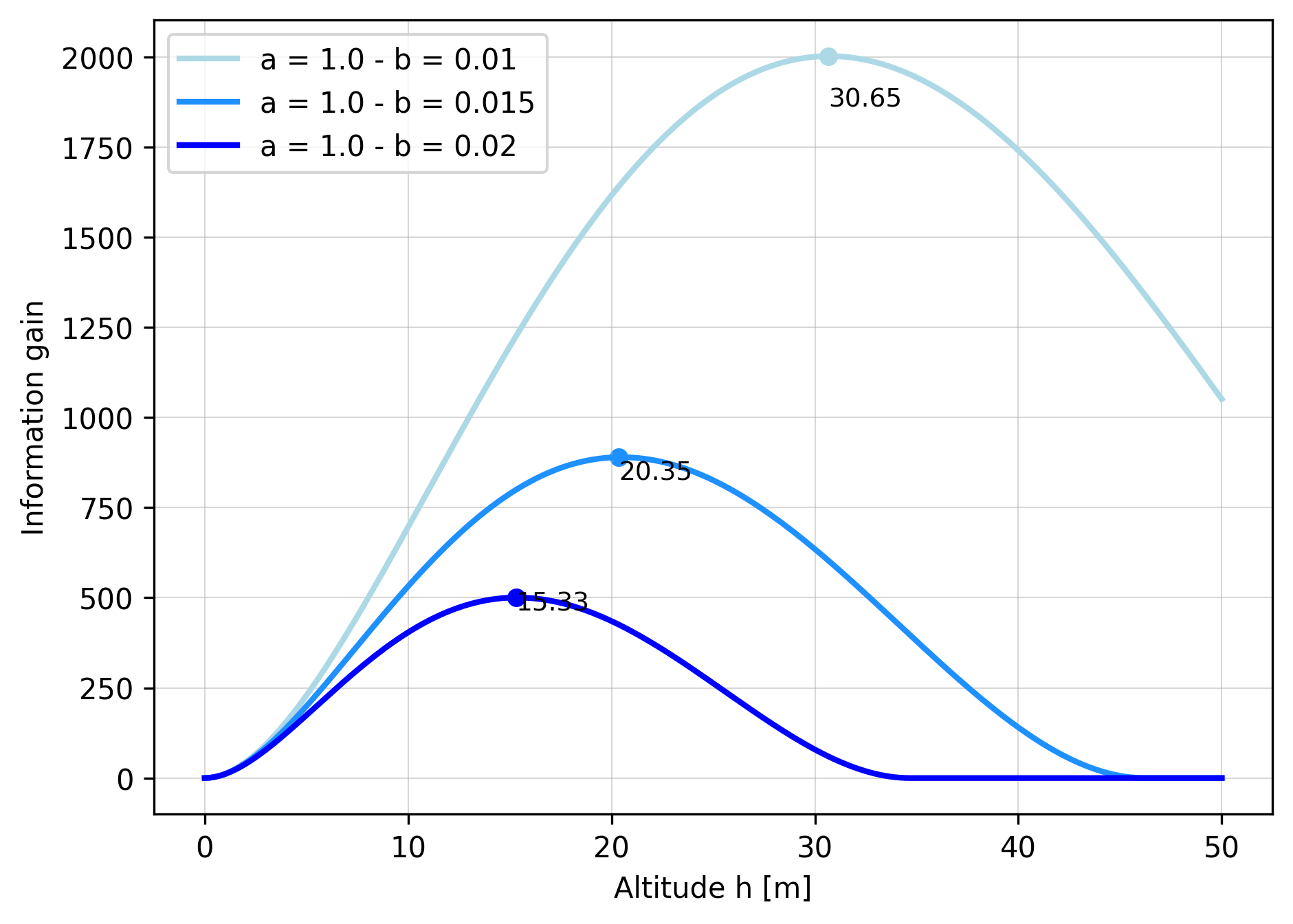}
        \label{fig:ig_altitude}
    }
    \caption{
    Altitude-dependent sensing analysis. Left: sensor uncertainty variation for different values of parameter \(b\). Right: information gain variation with altitude, where marked points indicate the optimal sensing altitude for each configuration.
    }
    \label{fig:altitude_sensor_analysis}
\end{figure}

To model altitude-dependent sensing uncertainty, we use the probabilistic sensor model defined in \cite{b4}:

\begin{equation}
p(z|m,x)=
\begin{cases}
1-\sigma(x), & z=m \\
\sigma(x), & z\neq m
\end{cases}
\label{eq:sensor_model}
\end{equation}

where \(m \in \{0,1\}\) is the true terrain state, \(z \in \{0,1\}\) is the sensor observation, and \(x=\langle x,y,h\rangle\) denotes the UAV pose. The term \(\sigma(x)\) represents the probability of a sensing error, assuming symmetric false positive and false negative rates. Since image quality decreases with altitude, the sensing uncertainty is modeled as

\begin{equation}
\sigma(x)=a\left(1-e^{-bh}\right),
\qquad a,b>0
\label{eq:sigma}
\end{equation}

where \(\sigma(x)\) is upper bounded by \(0.5\), corresponding to a completely uninformative sensor. Fig.~\ref{fig:altitude_sensor_analysis} illustrates the effect of altitude on sensing uncertainty and information gain.

\subsection{Factor Graph Terrain Mapping}

Terrain mapping is formulated as a probabilistic estimation problem over a static binary environment. Each terrain cell is represented by a random variable \(m_i\), and the belief map \(B\) is recursively updated as UAV observations are collected. To account for the spatial structure commonly observed in agricultural fields, neighboring terrain cells are connected using a pairwise Conditional Random Field (CRF).

The CRF represents the terrain as an undirected graph, where the conditional distribution over hidden terrain states and observations is factorized as

\begin{equation}
p(m|s)=
\frac{1}{Z(s)}
\prod_{i=1}^{k}\phi_e(m_i,o)
\prod_{\substack{i,j\in\{1,\ldots,k\}\\(i,j)\in E}}
\phi_p(m_i,m_j,o)
\label{eq:crf}
\end{equation}

where \(m=(m_1,\ldots,m_n)\in M\) denotes the set of terrain cells, \(s=(o_1,\ldots,o_T)\in S\) is the observation history, \(Z(s)\) is the partition function, and \(E\) is the set of neighboring cell connections. Each observation contains the UAV position and the sensor measurements inside the footprint:

\begin{equation}
o_t=(x_t,z_{t,1},\ldots,z_{t,k})
\label{eq:observation}
\end{equation}

with \(k=|C(x_t)|\). The evidence factor relates each hidden cell state to the corresponding sensor observation:

\begin{equation}
\phi_e(m_i,o)=p(z_i|m_i,x)
\label{eq:evidence}
\end{equation}

while the pairwise factor models the spatial dependency between neighboring cells:

\begin{equation}
\phi_p(m_i,m_j,o)=w_{m_i,m_j}(o)
\label{eq:pairwise}
\end{equation}

where \(w_{m_i,m_j}(o)\) controls the strength of the spatial correlation between adjacent cells. At each step, the posterior marginal probability \(p(m_i|o_{1:t})\) is estimated for the observed cells, and the belief map is updated accordingly. Inference over the CRF is performed using the Sum-Product variant of Loopy Belief Propagation \cite{b10}.

\subsubsection{Pairwise Factor Weights}

The pairwise factor \(w_{m_i,m_j}(o)\) controls how strongly neighboring cells influence each other during belief propagation. In this work, various types of pairwise weighting strategies are considered: equal, biased, and adaptive weights.

In the equal weighting strategy, all neighboring configurations are assigned the same value:

\begin{equation}
w_{m_i,m_j}(o)=0.5,
\qquad
\forall \; m_i,m_j \in \{0,1\}.
\label{eq:equal_weight}
\end{equation}

This represents the case where no prior spatial preference is imposed between adjacent cells. In the biased weighting strategy, neighboring cells with the same state are encouraged by assigning a larger weight to equal configurations:

\begin{equation}
w_{m_i,m_j}(o)=
\begin{cases}
0.7, & \text{if } m_i=m_j, \\
0.3, & \text{if } m_i \neq m_j.
\end{cases}
\label{eq:biased_weight}
\end{equation}

The adaptive weighting strategy estimates local spatial correlation directly from the observations collected during the mission. For each observed footprint, a set of non-overlapping Von Neumann neighborhoods is extracted. Each sample is represented by the pair \((c_i,n_i)\), where \(c_i \in \{0,1\}\) is the central cell value and \(n_i \in \{0,1,2,3,4\}\) is the number of positive four-neighbor cells. The sample set is denoted as \(D=\{(c_1,n_1),\ldots,(c_h,n_h)\}\), where \(|D|=|C(x)|/9\).

The local Pearson correlation coefficient is estimated as

\begin{equation}
\hat{\rho}_{c,n}=
\frac{
\sum_{i=1}^{h}(c_i-\bar{c})(n_i-\bar{n})
}{
\sqrt{\sum_{i=1}^{h}(c_i-\bar{c})^2}
\sqrt{\sum_{i=1}^{h}(n_i-\bar{n})^2}
},
\label{eq:pearson_weight}
\end{equation}

where \(\bar{c}\) and \(\bar{n}\) are the mean values of \(c_i\) and \(n_i\) over the sample set \(D\). Since \(\hat{\rho}_{c,n}\in[-1,1]\), a normalization function \(f(\cdot)\) is used to map the correlation value into \([0,1]\). The adaptive pairwise weight is then defined as

\begin{equation}
w_{m_i,m_j}(z)=
\begin{cases}
f(\hat{\rho}_{c,n}), & \text{if } m_i=m_j, \\
1-f(\hat{\rho}_{c,n}), & \text{if } m_i \neq m_j.
\end{cases}
\label{eq:adaptive_weight}
\end{equation}

In the experiments, several adaptive normalization functions are considered. Since the Pearson correlation coefficient satisfies \(\hat{\rho}_{c,n}\in[-1,1]\), normalization functions are used to transform the correlation value into the interval \([0,1]\).

The \textbf{standard sigmoid normalization} is defined as

\begin{equation}
f_{\mathrm{sig}}(\hat{\rho}_{c,n})=
\frac{1}{1+e^{-\hat{\rho}_{c,n}}},
\label{eq:sigmoid_weight}
\end{equation}

while the \textbf{soft sigmoid normalization} introduces a smoothing factor \(k\):

\begin{equation}
f_{\mathrm{soft\_sig}}(\hat{\rho}_{c,n})=
\frac{1}{1+e^{-k\hat{\rho}_{c,n}}},
\label{eq:soft_sigmoid_weight}
\end{equation}

where \(k=0.5\). The \textbf{hyperbolic tangent normalization} is expressed as

\begin{equation}
f_{\mathrm{tanh}}(\hat{\rho}_{c,n})=
\frac{\tanh(\hat{\rho}_{c,n})+1}{2},
\label{eq:tanh_weight}
\end{equation}

and its softer variant, the \textbf{soft hyperbolic tangent normalization}, is defined as

\begin{equation}
f_{\mathrm{soft\_tanh}}(\hat{\rho}_{c,n})=
\frac{\tanh(k\hat{\rho}_{c,n})+1}{2},
\label{eq:soft_tanh_weight}
\end{equation}

with \(k=0.5\). For computationally simpler approximations, the \textbf{hard sigmoid normalization} is given by

\begin{equation}
f_{\mathrm{hard\_sig}}(\hat{\rho}_{c,n})=
\mathrm{clip}
\left(
\frac{\hat{\rho}_{c,n}+1}{2},
0,1
\right),
\label{eq:hard_sigmoid_weight}
\end{equation}

whereas the \textbf{hard tanh normalization} is defined as

\begin{equation}
f_{\mathrm{hard\_tanh}}(\hat{\rho}_{c,n})=
\frac{
\mathrm{clip}(\hat{\rho}_{c,n},-1,1)+1
}{2}.
\label{eq:hard_tanh_weight}
\end{equation}

When the estimated spatial correlation is weak, the adaptive weights remain close to \(0.5\). As the correlation increases, the weights become more biased toward locally coherent neighboring cell configurations.

\subsection{Information-Gain Planning}

The UAV planning policy selects actions that maximize the expected reduction of uncertainty in the belief map. At each decision step, the UAV evaluates the admissible action set \(A=\{\text{up, down, front, back, left, right, hover}\}\) and selects the action \(a \in A(x_t)\) that provides the highest expected information gain over a one-step planning horizon. Actions that move the UAV outside the admissible configuration space are discarded. For a candidate future position \(\hat{x}_{t+1}\), the information gain associated with a terrain cell \(m_i\) is defined as

\begin{equation}
IG(m_i,\hat{x}_{t+1}) =
H(m_i)-H(m_i|\hat{o}_{t+1})
\label{eq:ig_cell}
\end{equation}

where \(H(m_i)\) is the current entropy of the cell belief:

\begin{equation}
H(m_i)=
-\sum_{m_i}
p^{-}(m_i)\log(p^{-}(m_i))
\label{eq:entropy_prior}
\end{equation}

and \(H(m_i|\hat{o}_{t+1})\) is the expected posterior entropy after a possible future observation:

\begin{equation}
H(m_i|\hat{o}_{t+1})=
\sum_{\hat{z}_{t+1,i}}
p(\hat{o}_{t+1})
\left[
-\sum_{m_i}
p^{+}(m_i)\log(p^{+}(m_i))
\right]
\label{eq:entropy_posterior}
\end{equation}

with

\begin{equation}
p^{-}(m_i)=
p(m_i|z_{0:t,i},x_{0:t})
\label{eq:prior_belief}
\end{equation}

\begin{equation}
p(\hat{o}_{t+1})=
p(\hat{z}_{t+1,i},\hat{x}_{t+1})
\label{eq:future_observation}
\end{equation}

\begin{equation}
p^{+}(m_i)=
p(m_i|z_{0:t,i},x_{0:t},\hat{z}_{t+1,i},\hat{x}_{t+1})
\label{eq:posterior_belief}
\end{equation}

The total information gain of an admissible action is computed by summing the cell-wise information gain over the future sensing footprint:

\begin{equation}
IG_a=
\sum_{m_i \in C(\hat{x}_{t+1})}
IG(m_i,\hat{x}_{t+1})
\label{eq:ig_action}
\end{equation}

The selected action is then

\begin{equation}
a^{*}=
\arg\max_{a\in A(x_t)}
IG_a .
\label{eq:best_action}
\end{equation}

This formulation naturally balances sensing accuracy and footprint size. Lower altitudes provide more accurate observations, while higher altitudes allow more cells to contribute to the total information gain. As shown in Fig.~\ref{}, this trade-off leads to a non-linear relation between altitude and information gain, with an intermediate altitude providing the maximum expected utility.

\subsection{Multi-UAV Information Sharing}

In multi-UAV scenarios, agents operate simultaneously and exchange information within a communication radius \(R\). This allows neighboring UAVs to reduce redundant observations and improve consistency between their local belief maps. To simulate asynchronous decision making, only one UAV is updated at each simulation step, and the update order among agents is randomized.

\subsubsection{Position Sharing}

In position sharing, each UAV broadcasts its current position to neighboring agents. The expected information gain of each candidate action is then discounted according to the overlap between sensing footprints. The overlap penalty between UAVs \(i\) and \(j\) is computed using the Intersection over Union (IoU):

\begin{equation}
\alpha_{ij}
=
1-
IoU(fp(x_i),fp(x_j))
\label{eq:iou_discount}
\end{equation}

where \(\alpha_{ij}\in[0,1]\) is a discount factor. Larger footprint overlap produces a stronger penalty, while non-overlapping footprints result in values close to one. The discounted information gain for UAV \(i\) is

\begin{equation}
IG_a^d
=
IG_a
\prod_{j\in N}
\alpha_{ij}
\label{eq:discounted_ig}
\end{equation}

where \(N\) is the set of neighboring UAVs within communication range. The action with the highest discounted information gain is then selected, encouraging UAVs to explore different regions of the field.

\subsubsection{Belief Sharing}

In belief sharing, UAVs exchange newly acquired belief information to improve the consistency of their local maps. Each UAV maintains an auxiliary belief map \(\hat{B}\), called the \textit{news belief}, which stores only the observations collected since the last communication event. Exchanging only new information reduces repeated incorporation of the same observations and helps limit double counting.

When a UAV receives a news belief \(\hat{B}\) from a neighboring agent, it fuses it with its local belief map \(B\). In this work, different probabilistic fusion rules are compared. The original Bayesian fusion rule, motivated by probabilistic inference principles \cite{b15}, is

\begin{equation}
B^{f}=
\frac{B\hat{B}}
{B\hat{B}+(1-B)(1-\hat{B})},
\label{eq:belief_fusion_bayesian}
\end{equation}

where \(B^{f}\) is the fused belief. This rule reinforces agreement between local and received beliefs. As simpler alternatives, \textbf{Weighted-average and consensus fusion} are defined as

\begin{equation}
B^{f}=\alpha B+(1-\alpha)\hat{B},
\label{eq:belief_fusion_weighted_average}
\end{equation}

\begin{equation}
B^{f}=B+\epsilon(\hat{B}-B),
\label{eq:belief_fusion_consensus}
\end{equation}

where \(\alpha\) controls the trust assigned to the local belief and \(\epsilon\) controls the consensus update strength. We also evaluate \textbf{Log-odds-based fusion}. Defining

\begin{equation}
L(B)=\log\left(\frac{B}{1-B}\right),
\label{eq:log_odds}
\end{equation}

the log-odds fusion and \textbf{Covariance-intersection-inspired fusion} are written as

\begin{equation}
B^{f}=
\frac{1}
{1+\exp[-(L(B)+L(\hat{B}))]},
\label{eq:belief_fusion_log_odds}
\end{equation}

\begin{equation}
B^{f}=
\frac{1}
{1+\exp[-(\omega L(B)+(1-\omega)L(\hat{B}))]},
\label{eq:belief_fusion_ci}
\end{equation}

where \(\omega\in[0,1]\) controls the relative contribution of each belief source.

To account for altitude-dependent sensing reliability, \textbf{Confidence-weighted fusion} assigns larger weights to beliefs obtained from lower-altitude, more reliable observations:

\begin{equation}
w_{\mathrm{local}}=1-\sigma(h_{\mathrm{local}}), \qquad
w_{\mathrm{sender}}=1-\sigma(h_{\mathrm{sender}}),
\label{eq:confidence_weights}
\end{equation}

\begin{equation}
B^{f}=
\frac{
w_{\mathrm{local}}B+w_{\mathrm{sender}}\hat{B}
}
{
w_{\mathrm{local}}+w_{\mathrm{sender}}
}.
\label{eq:belief_fusion_confidence}
\end{equation}

Finally, a \textbf{Dempster--Shafer-style binary fusion} rule is tested by defining the conflict

\begin{equation}
K=B(1-\hat{B})+(1-B)\hat{B},
\label{eq:ds_conflict}
\end{equation}

and computing

\begin{equation}
B^{f}=
\frac{B\hat{B}}{1-K}.
\label{eq:belief_fusion_ds}
\end{equation}

In all cases, probabilities are clipped to remain within \((0,1)\) to avoid numerical instability. After fusion, \(B^{f}\) replaces the local belief map \(B\). In the implementation, separate news beliefs can be maintained for different neighboring UAVs, so that each communication event resets only the information exchanged with the corresponding agent.

\section{Experimental Setup}

\begin{figure}[!t]
    \centering
    \includegraphics[
        width=\columnwidth,height=5.4cm, keepaspectratio]{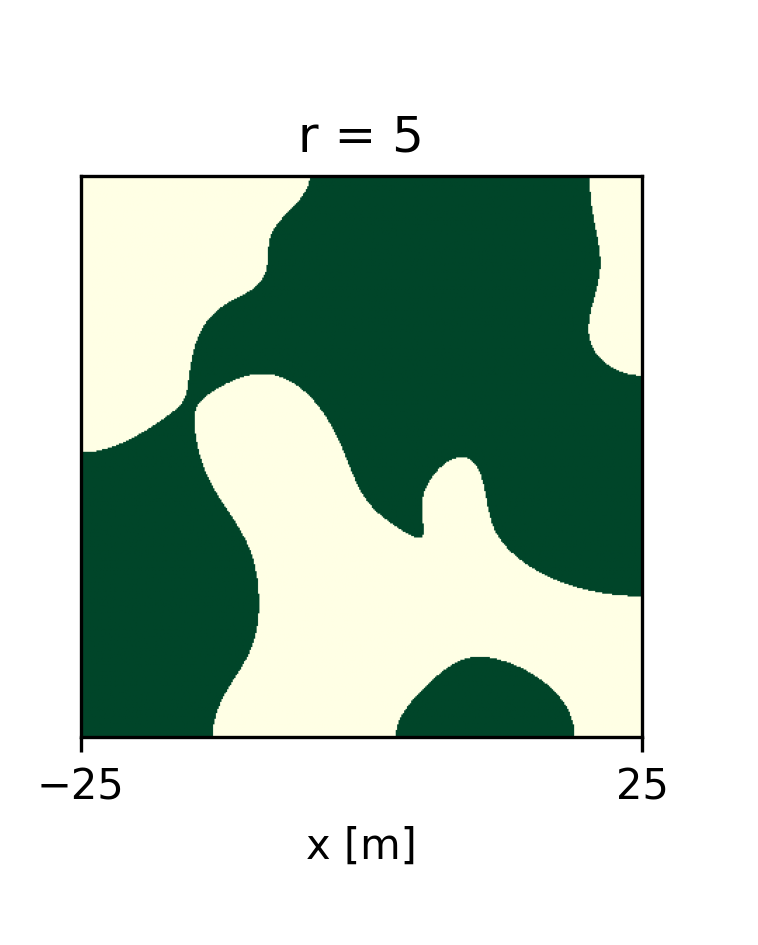}
    \caption{
    Examples of synthetically generated correlated Gaussian random field terrains for cluster radius value \(r=5\).
    }
    \label{fig:synthetic_terrains}
\end{figure}

The proposed framework is evaluated using both simulated correlated terrains and real UAV-derived agricultural imagery. In the simulated experiments, the monitored field covers an area of \(50\,\mathrm{m} \times 50\,\mathrm{m}\) and is discretized into \(400 \times 400\) cells, corresponding to a spatial resolution of \(12.5\,\mathrm{cm}\) per cell. Synthetic binary terrains are generated using correlated Gaussian random fields with different cluster radius values \(r \in \{0,1,2,3,4,5\}\). As shown in Fig.~\ref{fig:synthetic_terrains}, the parameter \(r\) controls the spatial correlation of the terrain: small values produce fragmented and weakly correlated patterns, while larger values produce smoother and larger homogeneous clusters. Thus, the figure illustrates changes in spatial structure rather than changes in standard binary entropy. For similar vegetation or non-vegetation proportions, the global class entropy may remain similar, while the spatial disorder decreases as \(r\) increases.

The UAV configuration space is discretized in three dimensions. In the horizontal plane, a resolution of \(\approx 3\,\mathrm{m}\) is used, resulting in \(17\) admissible waypoints along each side of the terrain. The altitude space is discretized into six levels, ranging from approximately \(5.4\,\mathrm{m}\) to \(32.4\,\mathrm{m}\), with a vertical resolution of approximately \(5.4\,\mathrm{m}\). Unless otherwise stated, the sensor uncertainty parameters are fixed to \(a=1\) and \(b=0.015\), and the false positive and false negative rates associated with the six altitude levels are set to \(\{0.07,0.15,0.22,0.28,0.33,0.40\}\). These values reflect the increasing sensing uncertainty at higher altitudes. The camera FoV is fixed to \(30^\circ\).

For real-world validation, a UAV orthomosaic image is extracted from the publicly available WeedMap dataset developed by ETH Zurich \cite{b11}. The dataset contains multispectral UAV imagery acquired over agricultural sugar beet fields using MicaSense RedEdge and Sequoia cameras. A large RGB orthomosaic from the RedEdge subset is used as the input image. Non-valid black background regions are removed using grayscale thresholding, and a square crop corresponding to approximately \(50 \times 50\,\mathrm{m}\) is extracted from the valid agricultural region. The crop is then resized to \(400 \times 400\) cells to match the simulated terrain resolution.

\begin{figure}[t]
    \centering
    \subfloat[Original UAV orthomosaic image.]{
        \includegraphics[
            width=0.44\columnwidth,
            height=4.2cm,
            keepaspectratio
        ]{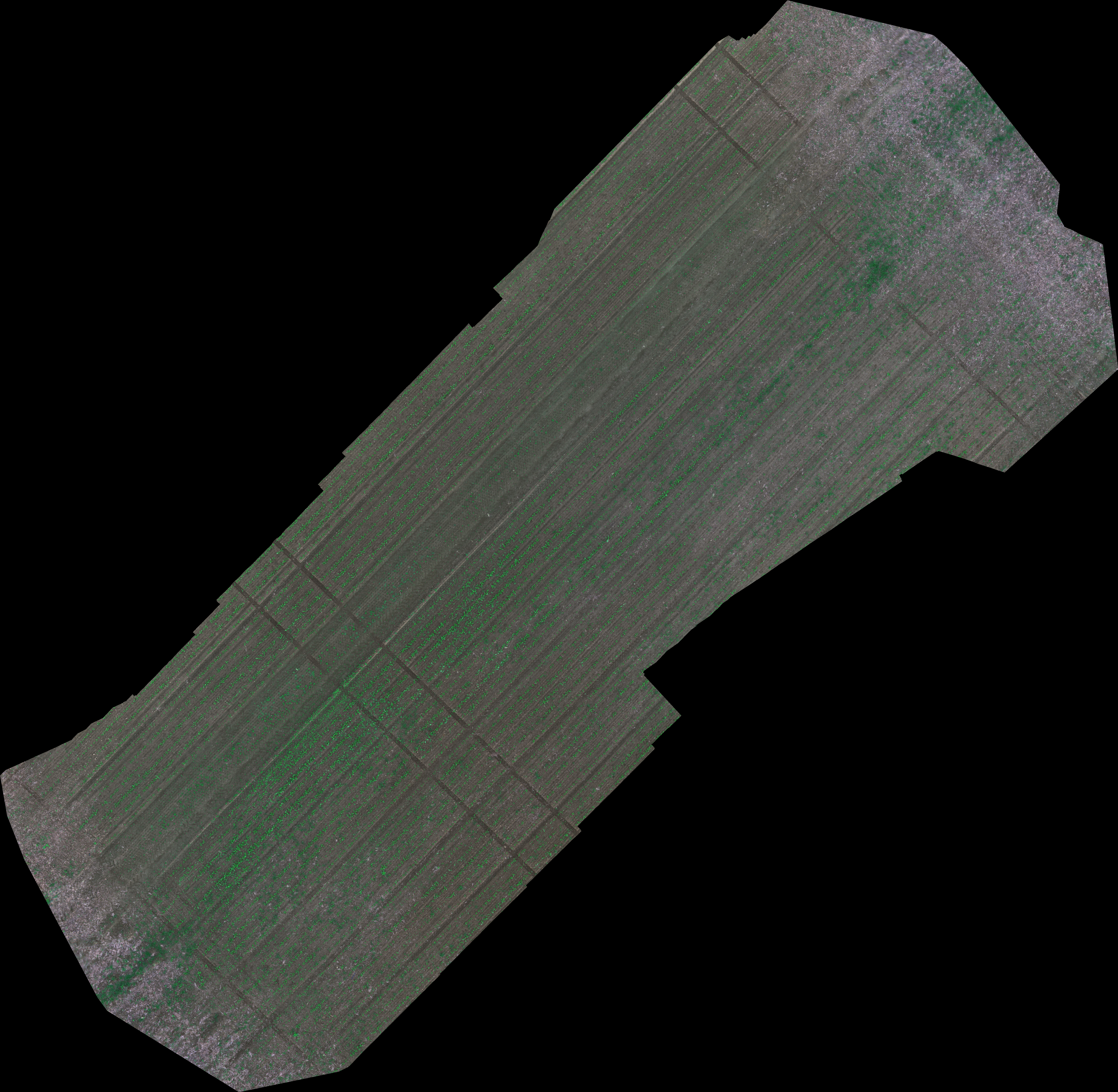}
        \label{fig:real_uav_rgb}
    }%
    \hfill
    \subfloat[Generated binary terrain map.]{
        \includegraphics[
            width=0.44\columnwidth,
            height=4.2cm,
            keepaspectratio
        ]{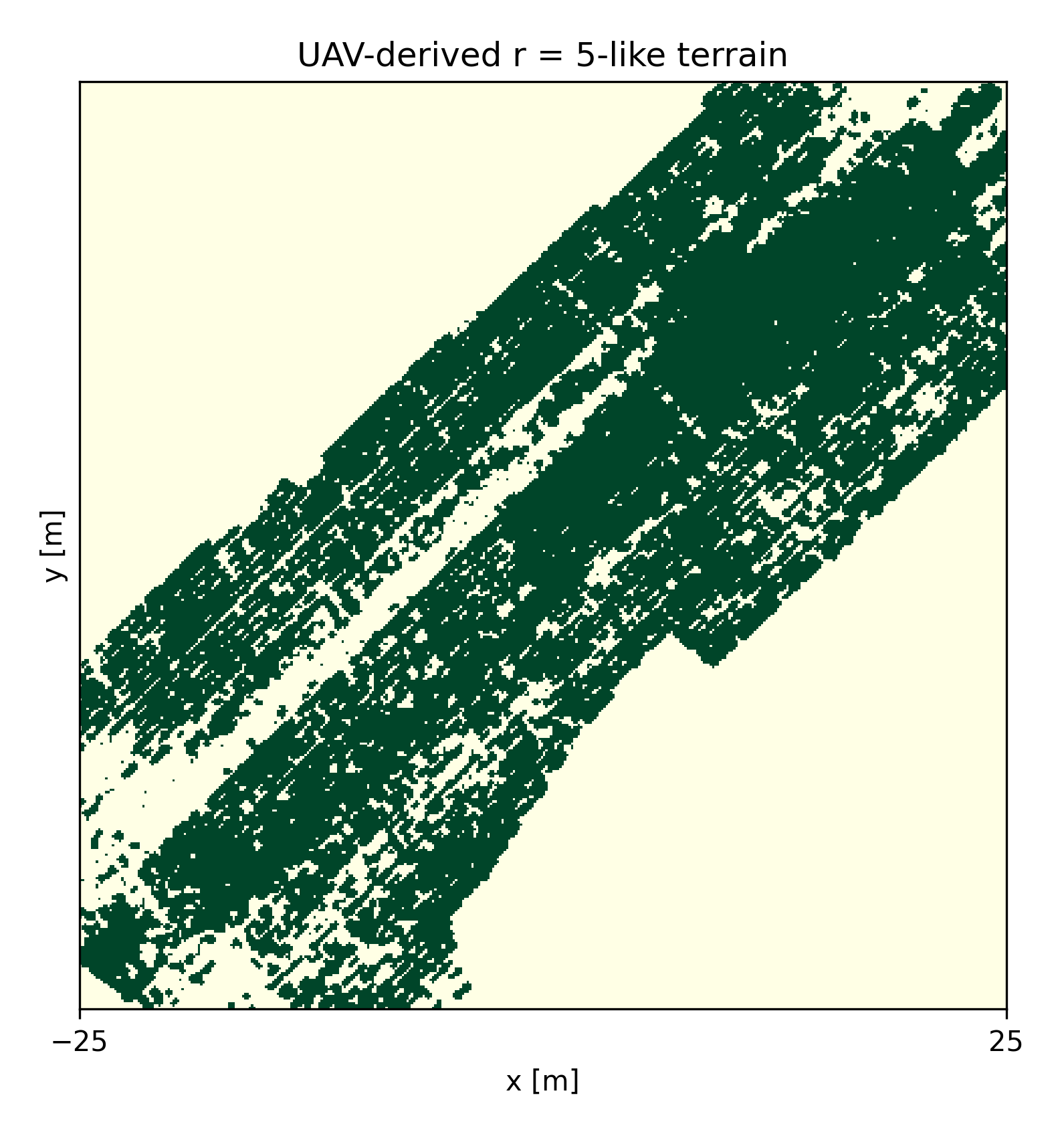}
        \label{fig:real_uav_binary}
    }
    \caption{
    Real UAV-based terrain generation process. Left: original UAV orthomosaic image captured over an agricultural field. Right: binary vegetation/no-vegetation terrain map used for real-world experimental validation.
    }
    \label{fig:real_world_terrain}
\end{figure}

Vegetation regions are identified using the Excess Green (ExG) index computed from normalized RGB channels:

\begin{equation}
ExG = 2G - R - B ,
\label{eq:exg}
\end{equation}

where \(R\), \(G\), and \(B\) denote the normalized red, green, and blue channels, respectively. Pixels with \(ExG > 0.08\) are classified as vegetation. The resulting binary map is smoothed using a Gaussian filter and thresholded to obtain a spatially correlated vegetation or no-vegetation terrain. The final terrain assigns vegetation cells to value \(1\) and no-vegetation cells to value \(0\). Fig.~\ref{fig:real_world_terrain} shows the original UAV orthomosaic and the generated binary terrain.

The evaluation considers four main quantities: total belief entropy, mean squared error (MSE), explored terrain coverage, and UAV altitude. Entropy measures the remaining uncertainty in the belief map, while MSE measures the difference between the estimated belief map and the ground-truth binary terrain. Coverage measures the percentage of terrain cells observed during the mission, and altitude is used to analyze the planner's sensing behavior.

The proposed Information Gain (IG) planner is compared with two baseline strategies: Random walk and Sweep. The Random planner selects actions uniformly at random, while the Sweep planner follows a predefined systematic back-and-forth coverage path over the field. For the spatial correlation analysis, equal, biased, and adaptive pairwise factor weights are compared. The adaptive weights are computed from the local Pearson correlation estimate and normalized using the sigmoid and hyperbolic tangent functions described in Section~II.

For the multi-UAV experiments, different team sizes \(N \in \{1,4\}\), communication ranges \(R \in \{5,25\}\,\mathrm{m}\), and information-sharing/fusion strategies are evaluated. The two communication ranges are selected to represent short-range and wider-range cooperation within the \(50\,\mathrm{m}\times50\,\mathrm{m}\) field. Although these values are not exact multiples of the \(\approx 3\,\mathrm{m}\) horizontal lattice resolution, they provide two clearly separated communication regimes for analyzing the effect of belief sharing. For each experimental condition, repeated runs are performed, and the reported curves are averaged over the runs.

\section{Results and Discussion}

\subsection{IG vs. Baseline Planners}

\begin{figure}[t]
    \centering
    \subfloat[Entropy.]{
        \includegraphics[width=0.47\columnwidth]{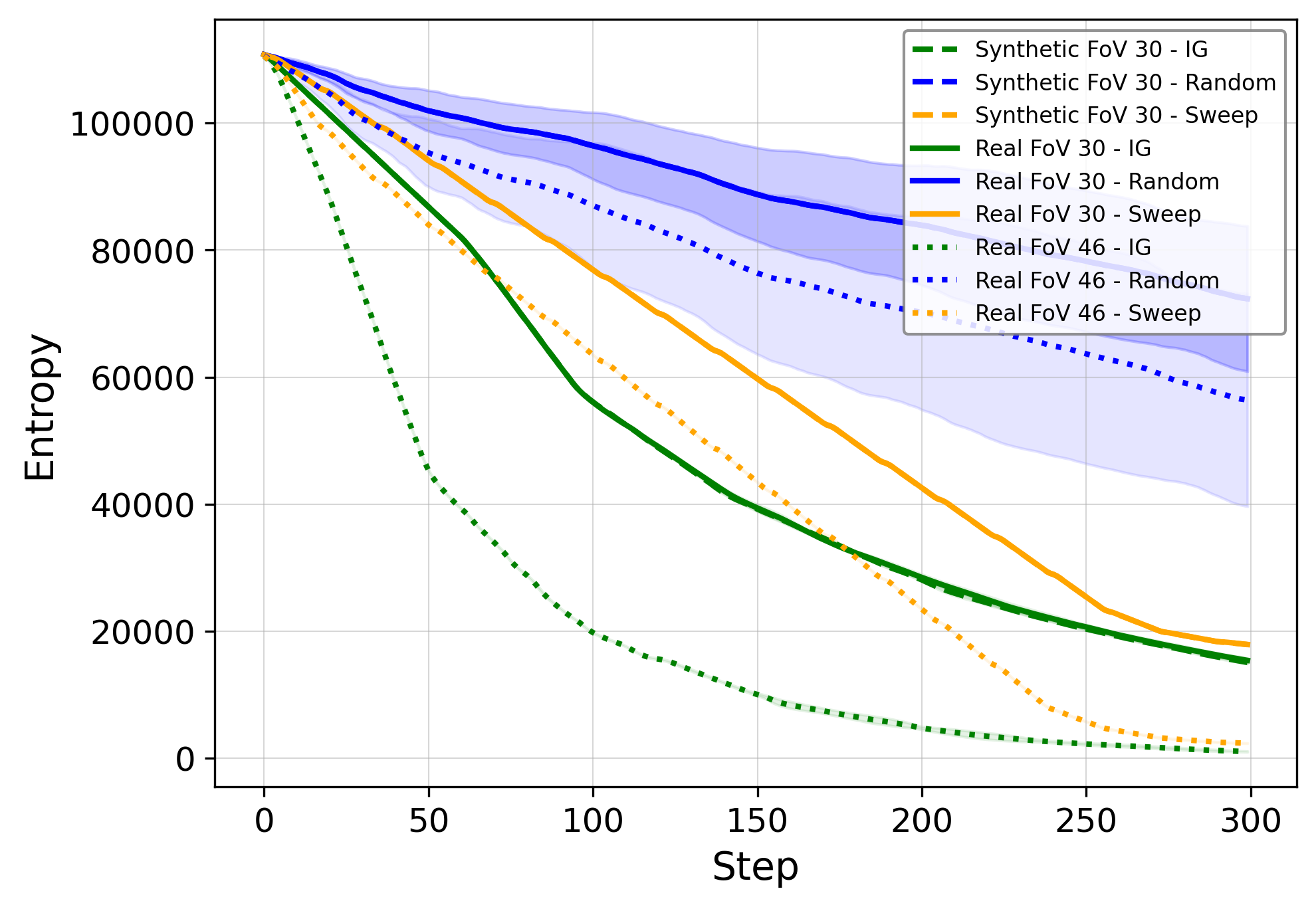}
        \label{fig:fig3_entropy}
    }
    \hfill
    \subfloat[MSE.]{
        \includegraphics[width=0.47\columnwidth]{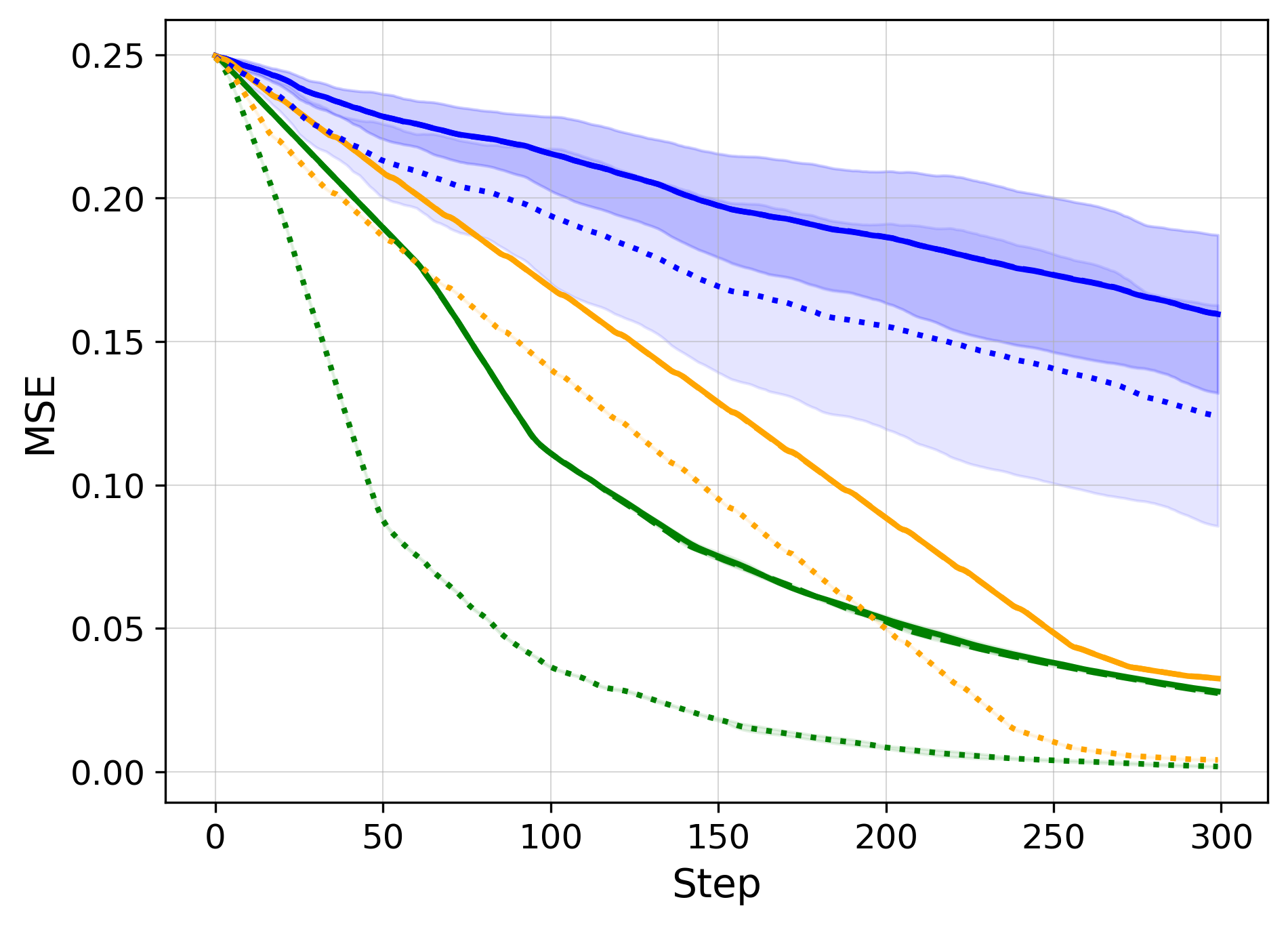}
        \label{fig:fig3_mse}
    }

    \vspace{0.1cm}

    \subfloat[Altitude.]{
        \includegraphics[width=0.47\columnwidth]{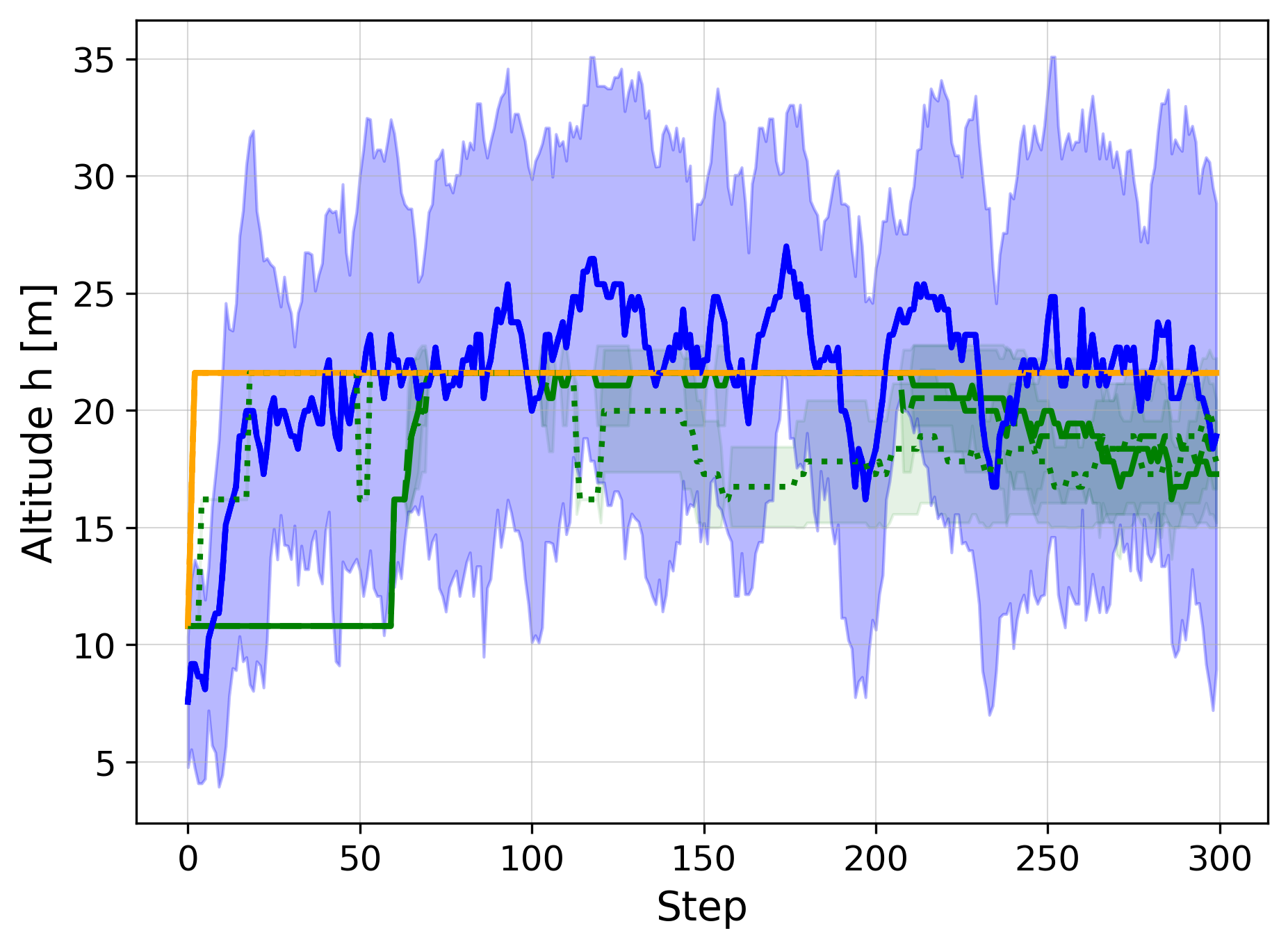}
        \label{fig:fig3_altitude}
    }
    \hfill
    \subfloat[Coverage.]{
        \includegraphics[width=0.47\columnwidth]{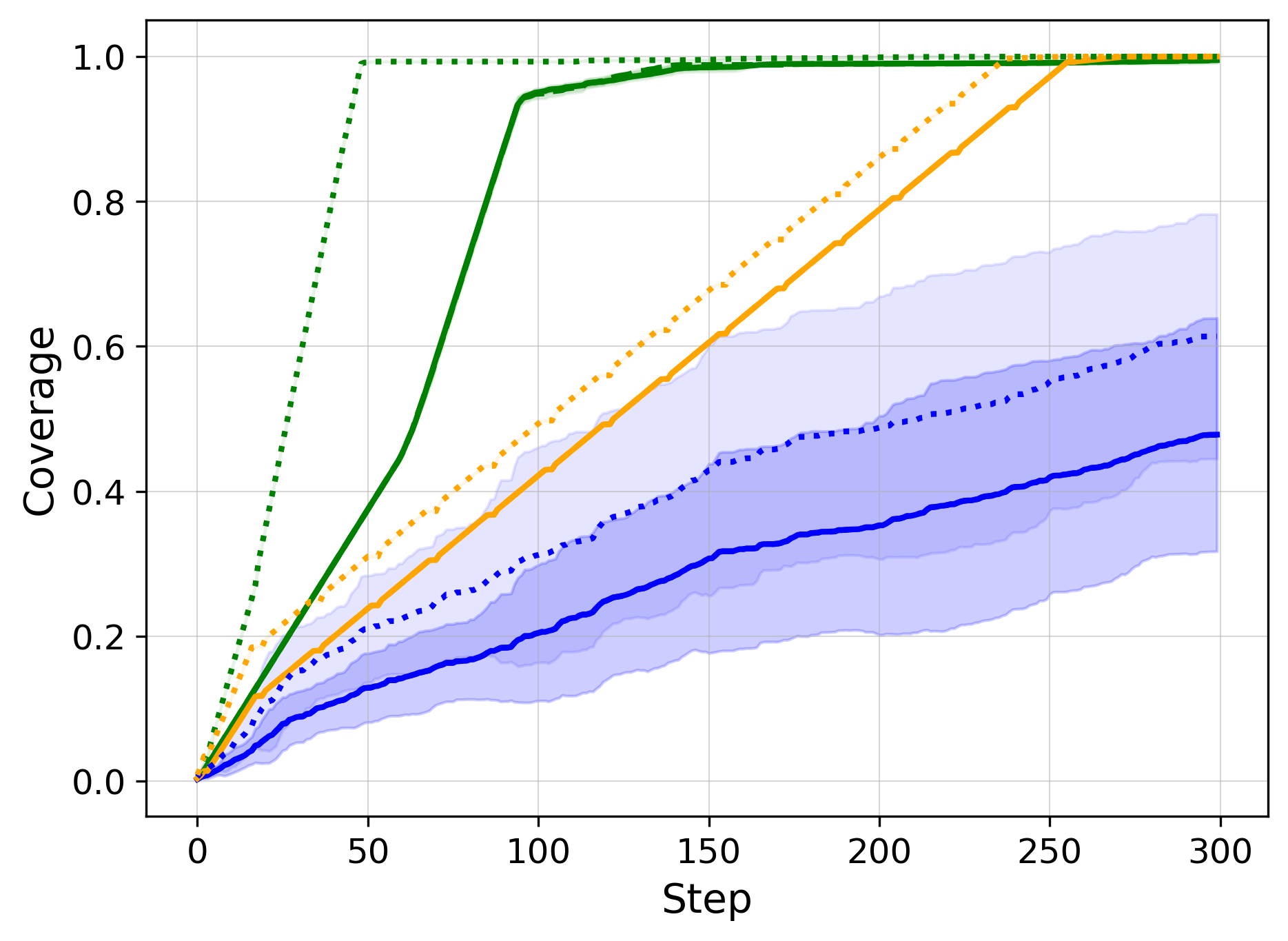}
        \label{fig:fig3_coverage}
    }

    \vspace{0.1cm}

    \subfloat[Real-world IG FoV \(46^\circ\).]{
        \includegraphics[width=0.47\columnwidth]{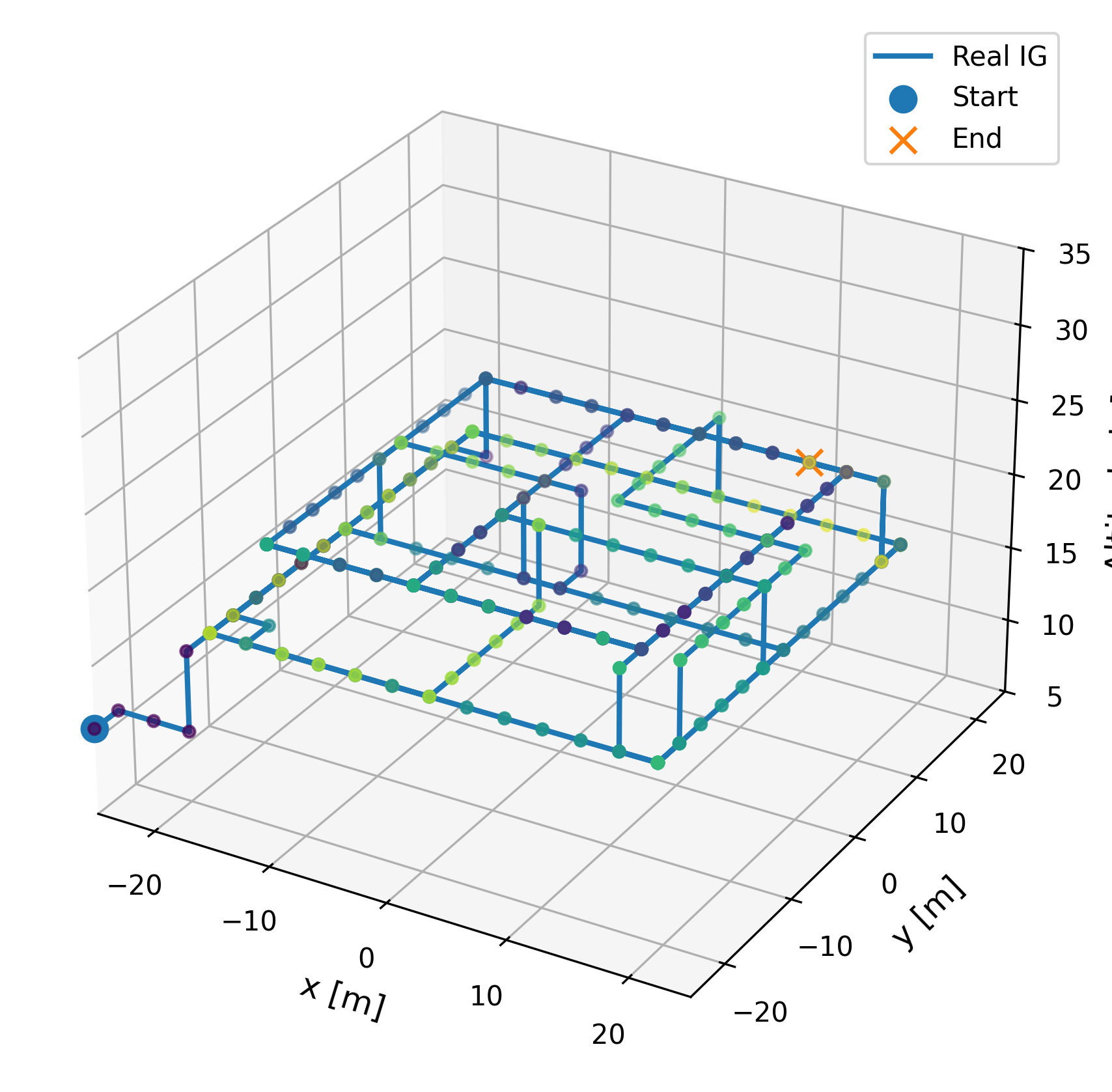}
        \label{fig:fig3_real_ig}
    }
    \hfill
    \subfloat[Real-world IG FoV \(30^\circ\).]{
        \includegraphics[width=0.47\columnwidth]{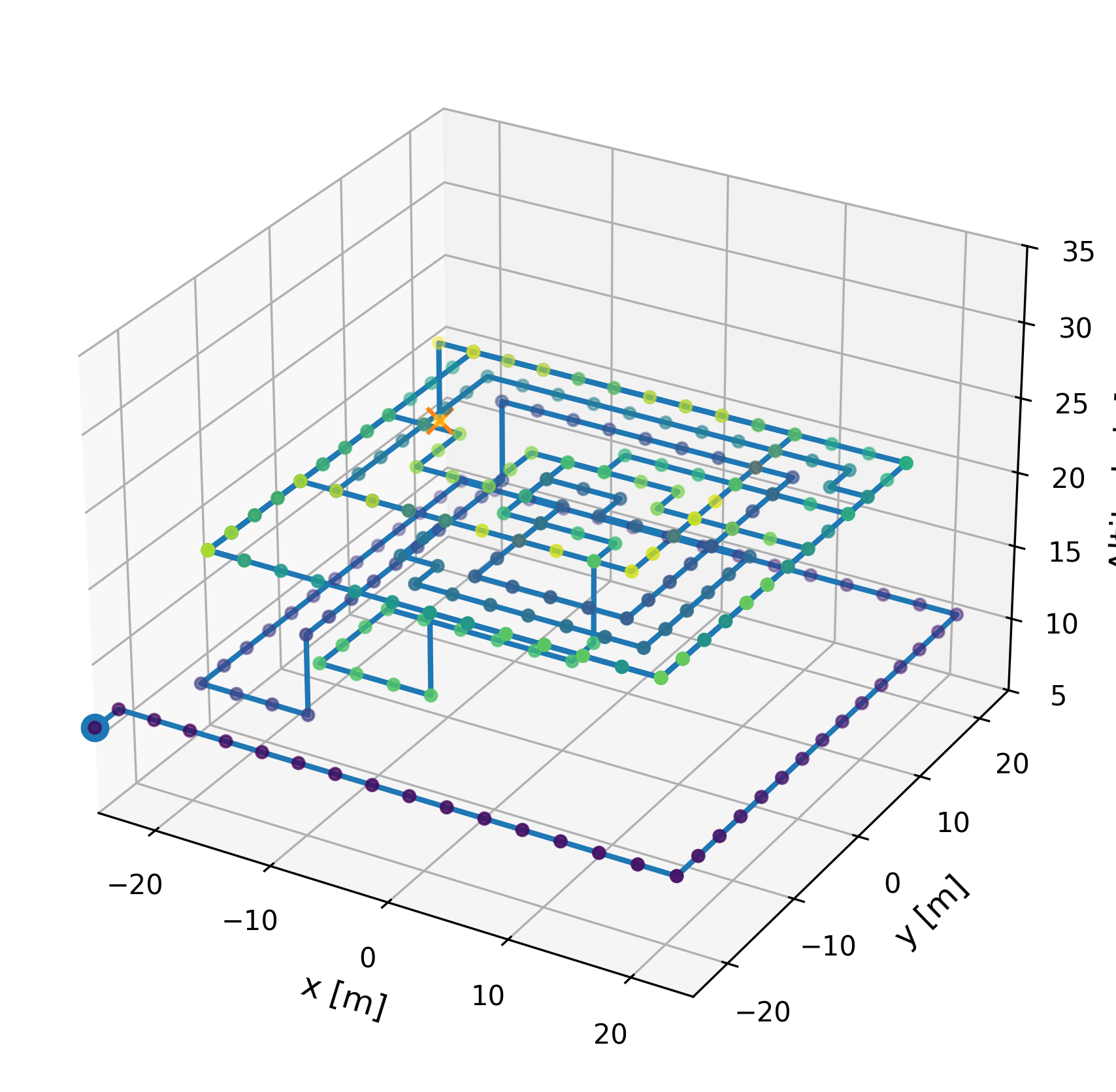}
        \label{fig:fig3_synthetic_ig}
    }

    \caption{
    Comparison of IG, Random, and Sweep planners in real-world and synthetic terrains. Top: entropy and MSE. Middle: UAV altitude and coverage. Bottom: 3D trajectories generated by the IG planner.
    }
    \label{fig:fig3_comparison}
\end{figure}

Fig.~\ref{fig:fig3_comparison} compares the proposed Information Gain (IG) planner with Random Walk and Sweep baseline strategies in synthetic and real-world terrains. The synthetic experiment was evaluated using a FoV of \(30^\circ\), while the real-world UAV-derived terrain was evaluated using both FoV \(30^\circ\) and FoV \(46^\circ\). The comparison considers entropy, MSE, altitude, coverage, and representative 3D trajectories. In the trajectory plots, Fig.~\ref{fig:fig3_real_ig} shows the real-world IG trajectory for FoV \(46^\circ\), while Fig.~\ref{fig:fig3_synthetic_ig} shows the real-world IG trajectory for FoV \(30^\circ\). Although each experiment consists of 300 decision cycles, the 3D trajectory does not necessarily display 300 visually distinguishable markers. Each marker represents a recorded UAV state \((x_t,y_t,h_t)\) at planning step \(t\), where \(x_t\) and \(y_t\) denote the horizontal grid position and \(h_t\) denotes the selected altitude. Since the UAV moves on a discrete 3D lattice, multiple decision steps may revisit the same or nearby grid states, causing markers to overlap in the visualization. Therefore, the number of visible points in Fig.~\ref{fig:fig3_real_ig} and Fig.~\ref{fig:fig3_synthetic_ig} can be smaller than the total number of executed planning steps. The blue marker indicates the initial state, and the orange cross indicates the final state at step 299.

For the synthetic terrain with FoV \(30^\circ\), IG achieves the strongest performance among the three planners over the 300-step planning horizon. In this experiment, the steps are indexed from \(t=0\) to \(t=299\), where each step corresponds to one UAV decision cycle: action selection, movement, sensing, and belief-map update. After 300 steps, IG reduces the entropy from approximately \(110747\) to \(15088\), corresponding to an entropy reduction of about \(86.38\%\). The MSE decreases from \(0.2496\) to \(0.0274\), corresponding to an MSE reduction of about \(89.02\%\). Sweep also performs well because it systematically covers the terrain, reducing the final MSE to \(0.0325\). However, IG achieves lower final entropy and MSE because it adapts its motion according to the current uncertainty of the belief map. Random Walk gives the weakest result, with a final MSE of approximately \(0.1596\), because its motion is not directed toward informative or uncovered regions. A similar trend is observed for the real UAV-derived terrain with FoV \(30^\circ\). Over the same 300 decision steps, IG reduces the entropy from approximately \(110747\) to \(15405\), and the MSE from \(0.2496\) to \(0.0280\). This corresponds to entropy and MSE reductions of about \(86.09\%\) and \(88.79\%\), respectively. Sweep reaches full coverage and obtains a final MSE of \(0.0325\), but IG still produces a slightly better final map estimate. Random Walk covers only about \(47.8\%\) of the terrain after 300 steps and remains with a much larger final MSE of \(0.1596\). These results show that the advantage of IG is not limited to synthetic terrain, but is also maintained on real UAV-derived agricultural imagery.

The effect of the camera FoV is clearly visible in the real-world experiments. With FoV \(46^\circ\), IG achieves a much stronger reduction in uncertainty and error than with FoV \(30^\circ\). After the 300-step horizon, the final entropy decreases to approximately \(1052\), and the final MSE decreases to \(0.0018\), corresponding to entropy and MSE reductions of about \(99.05\%\) and \(99.28\%\), respectively. This improvement occurs because the larger FoV increases the camera footprint, allowing the UAV to observe a larger portion of the terrain at each step while still adapting its altitude and position according to information gain. Sweep also benefits from the larger FoV and reaches a final MSE of \(0.0041\), but IG still gives the best final result. Random Walk improves compared with FoV \(30^\circ\), but its final MSE remains much higher at \(0.1242\), mainly because its exploration is not guided by map uncertainty. Ultimately, IG provides the best trade-off between coverage and map accuracy. Sweep is competitive because it guarantees systematic field coverage, but it does not adapt to the spatial distribution of uncertainty. Random Walk is useful as a lower-bound baseline, but it is inefficient for terrain mapping because it may repeatedly visit low-value regions while leaving other areas unexplored. The comparison between FoV \(30^\circ\) and FoV \(46^\circ\) further shows that sensing geometry strongly affects mapping performance: a wider FoV improves coverage and accelerates uncertainty reduction, especially when combined with an adaptive IG planner.

Table~\ref{tab:ig_baseline_final} summarizes the final planner performance after 300 steps. The results show that IG achieves the lowest final MSE in all evaluated settings. For the synthetic terrain with FoV \(30^\circ\), IG reaches a final MSE of 0.0274, compared with 0.0325 for Sweep and 0.1596 for Random Walk. A similar trend is observed in the real-world terrain with FoV \(30^\circ\). When the real-world FoV is increased to \(46^\circ\), IG further improves the final MSE to 0.0018, showing that a wider footprint combined with adaptive information-gain planning improves both coverage and map accuracy.

\begin{table}[!t]
\centering
\caption{Final performance of IG and baseline planners after 300 steps.}
\label{tab:ig_baseline_final}
\resizebox{\columnwidth}{!}{
\begin{tabular}{llccccc}
\hline
Terrain & FoV & Planner & Entropy & MSE & Coverage & Alt. \\
\hline
 & & IG     & 15087.74 & 0.0274 & 0.995 & 17.82 \\
Synthetic & \(30^\circ\) & Random & 72326.66 & 0.1596 & 0.478 & 18.90 \\
 & & Sweep  & 17898.26 & 0.0325 & 1.000 & 21.60 \\
\hline
 & & IG     & 15405.38 & 0.0280 & 0.995 & 17.28 \\
Real & \(30^\circ\) & Random & 72339.20 & 0.1595 & 0.478 & 18.90 \\
 & & Sweep  & 17911.50 & 0.0325 & 1.000 & 21.60 \\
\hline
 & & IG     & 1052.32  & 0.0018 & 1.000 & 19.44 \\
Real & \(46^\circ\) & Random & 56398.78 & 0.1242 & 0.614 & 18.90 \\
 & & Sweep  & 2369.21  & 0.0041 & 1.000 & 21.60 \\
\hline
\end{tabular}
}
\end{table}

\subsection{Effect of Spatial Correlation Weights}

\begin{figure}[t]
    \centering
    \subfloat[Entropy.]{
        \includegraphics[width=0.47\columnwidth]{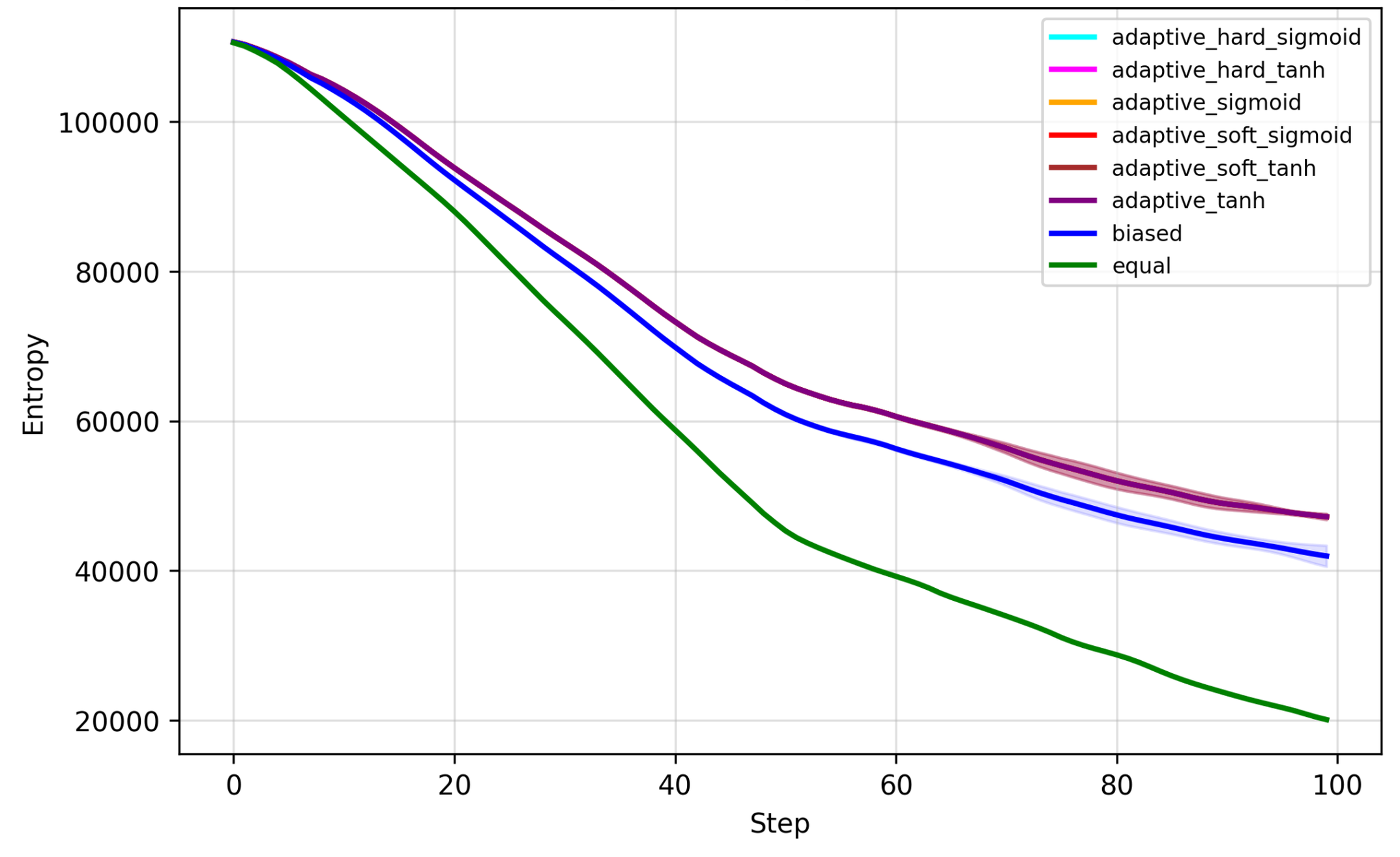}
        \label{fig:fig4_entropy}
    }
    \hfill
    \subfloat[MSE.]{
        \includegraphics[width=0.47\columnwidth]{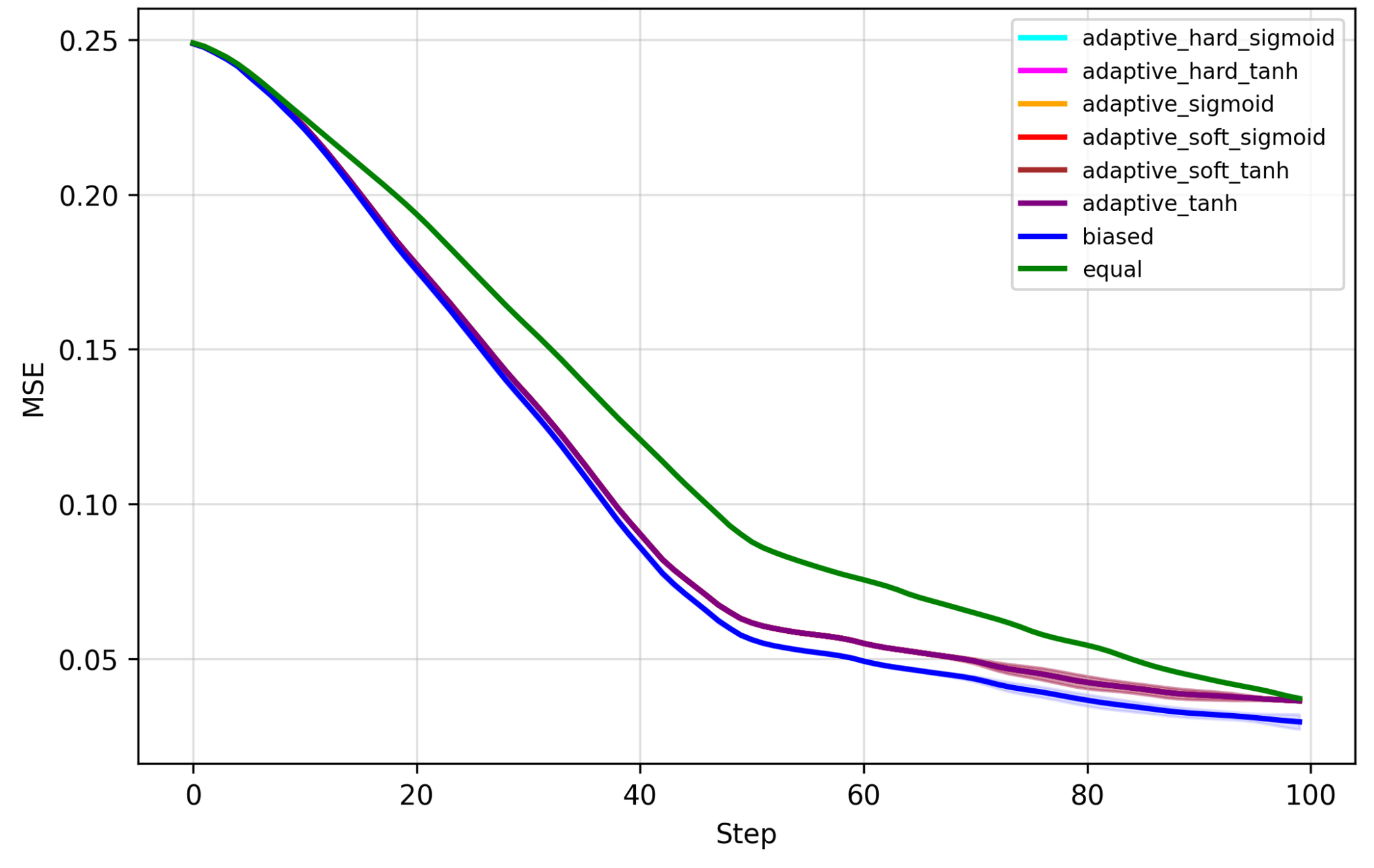}
        \label{fig:fig4_mse}
    }

    \vspace{0.1cm}

    \subfloat[Altitude.]{
        \includegraphics[width=0.47\columnwidth]{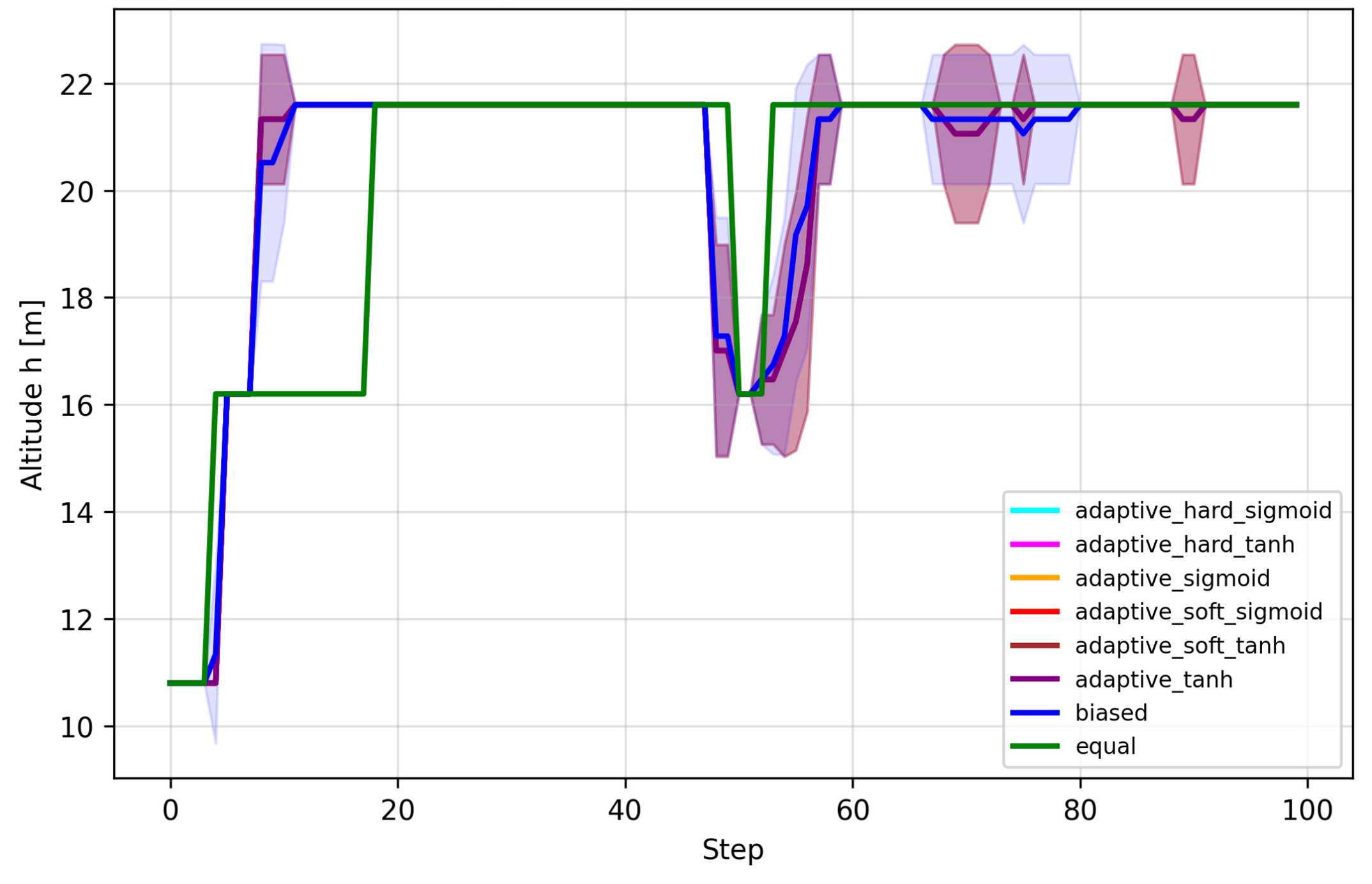}
        \label{fig:fig4_altitude}
    }
    \hfill
    \subfloat[Coverage.]{
        \includegraphics[width=0.47\columnwidth]{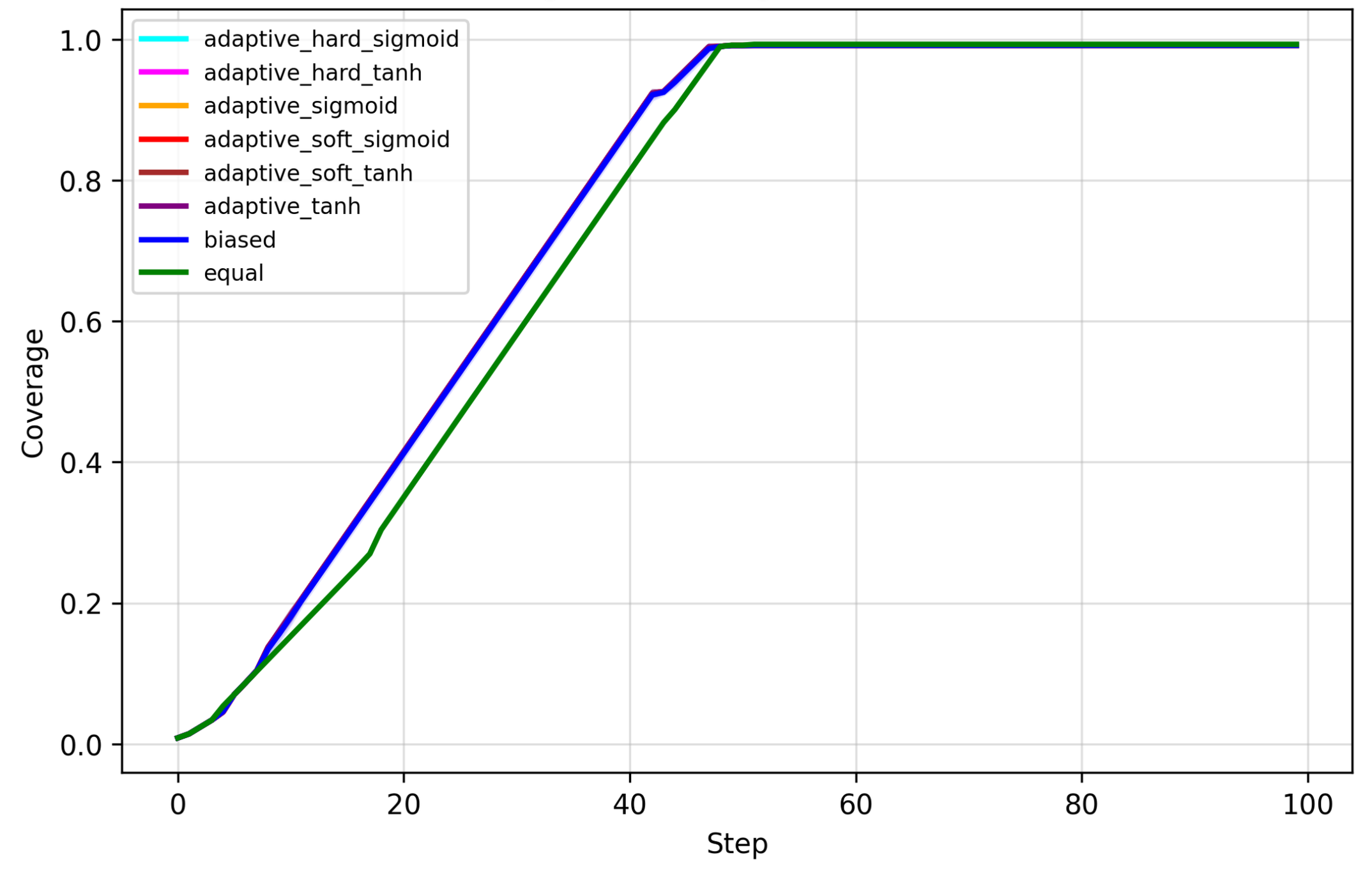}
        \label{fig:fig4_coverage}
    }

    \caption{
    Effect of different pairwise factor weights on real-world using the IG planner with a single UAV.
    }
    \label{fig:fig4_weights}
\end{figure}

This experiment evaluates the effect of pairwise spatial correlation weights on real-world UAV-derived terrain mapping. Only the IG planner is used, with a single UAV. The compared strategies are equal, biased, adaptive sigmoid, and adaptive hyperbolic tangent weights. Other adaptive variants are not included because their behaviour was very similar to the standard adaptive sigmoid and tanh cases and did not provide additional insight. Fig.~\ref{fig:fig4_weights} shows the evolution of entropy, MSE, coverage, and UAV altitude for the different weighting strategies. All methods start from a similar initial condition, with an MSE close to \(0.249\), entropy close to \(1.1\times 10^5\), and an initial altitude of approximately \(10.8\,\mathrm{m}\). During the first part of the mission, the UAV increases its altitude to expand the sensing footprint and rapidly explore the terrain. Around the middle of the mission, the altitude temporarily decreases to about \(16.2\,\mathrm{m}\), indicating a local refinement phase, before returning to higher-altitude sensing. The coverage curves show that all methods reach almost full coverage by approximately step 50, with final coverage values close to \(0.99\). Therefore, the main performance differences are mainly caused by the belief update strategy rather than by coverage.

The entropy results show that the equal-weight strategy gives the strongest uncertainty reduction. It decreases the entropy from approximately \(1.1\times 10^5\) to about \(2.0\times 10^4\), while the biased strategy ends around \(4.2\times 10^4\), and the adaptive sigmoid and adaptive tanh strategies end around \(4.7\times 10^4\). This indicates that the equal-weight model reduces uncertainty more strongly than the correlation-based models. One reason is that equal weights do not impose a strong spatial prior between neighbouring cells. Therefore, the belief update is mainly driven by direct UAV observations. This is useful in real UAV-derived agricultural imagery, where vegetation boundaries are irregular and local spatial correlations may change across the field. The MSE results show that all strategies reduce the initial error significantly. The MSE decreases from approximately \(0.249\) to about \(0.036\)--\(0.037\) for the equal, adaptive sigmoid, and adaptive tanh strategies, while the biased strategy reaches the lowest final MSE, around \(0.029\). Although the biased strategy gives the best final MSE, the equal strategy remains more stable in terms of uncertainty reduction and gives comparable mapping accuracy. The adaptive sigmoid and adaptive tanh curves are almost identical, showing that the choice between these two normalization functions has little effect in this experiment.

\begin{figure}[t]
    \centering
    \subfloat[IG, no sharing.]{
        \includegraphics[width=0.47\columnwidth]{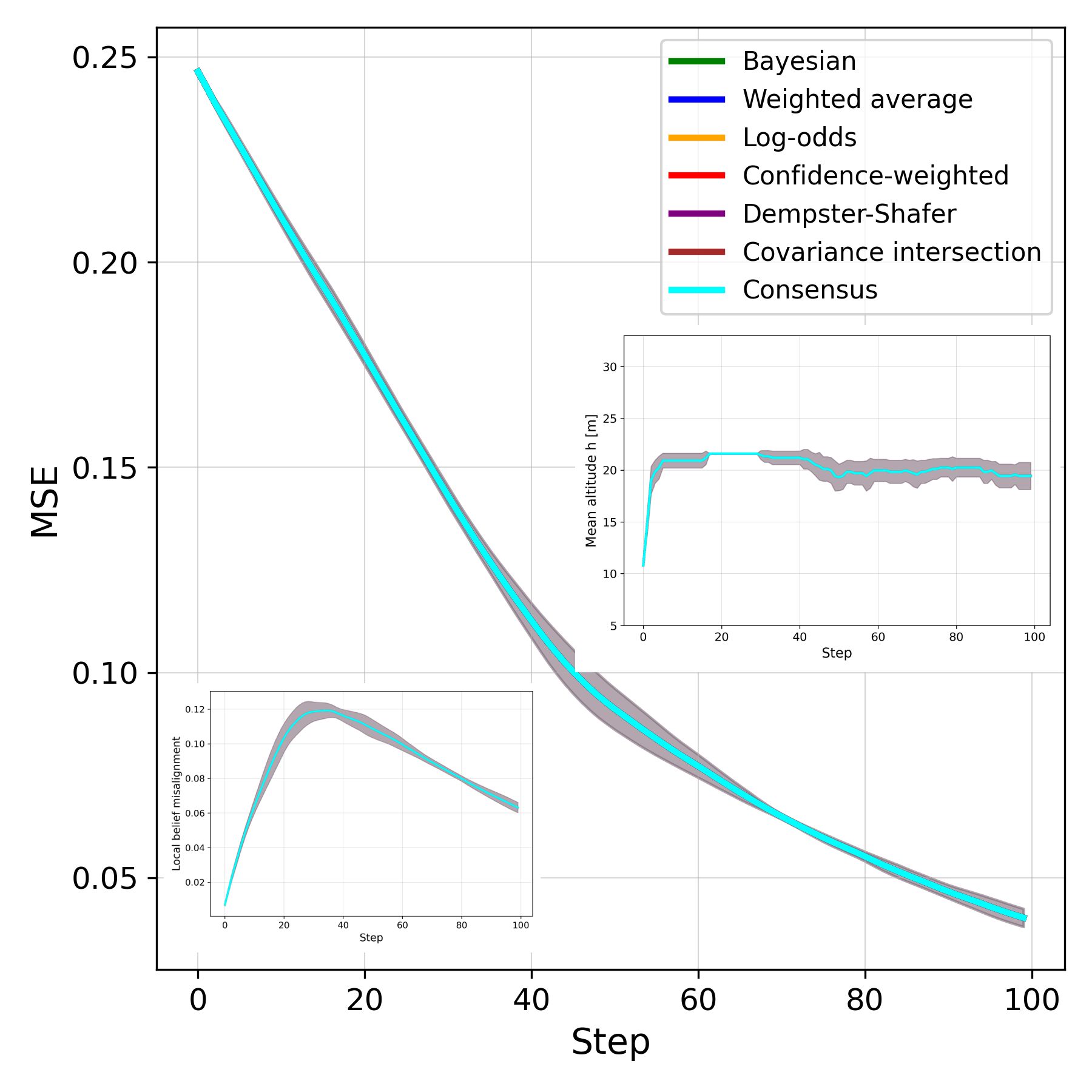}
        \label{fig:fig5_ig_none}
    }
    \hfill
    \subfloat[IGd, no sharing.]{
        \includegraphics[width=0.47\columnwidth]{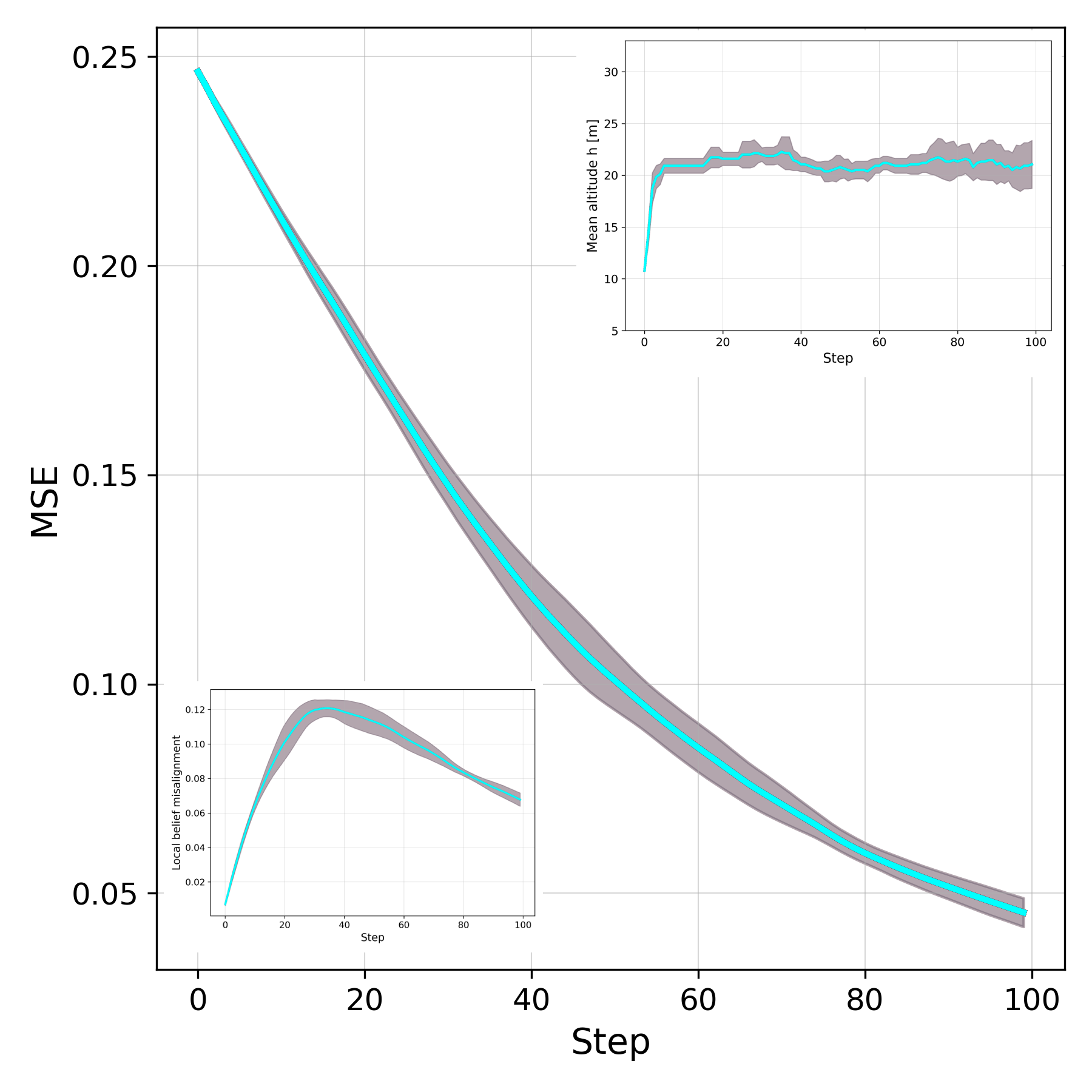}
        \label{fig:fig5_igd_none}
    }
    \caption{
    Comparison of different belief-fusion strategies for multi-UAV terrain mapping with \(N=4\) and no sharing. The main plots show the MSE evolution, while the inset plots show the corresponding mean altitude and local belief misalignment.
    }
    \label{fig:fig5_fusion_methods_none}
\end{figure}

\begin{figure}[t]
    \centering
    \subfloat[IG, BS, R=5.]{
        \includegraphics[width=0.47\columnwidth]{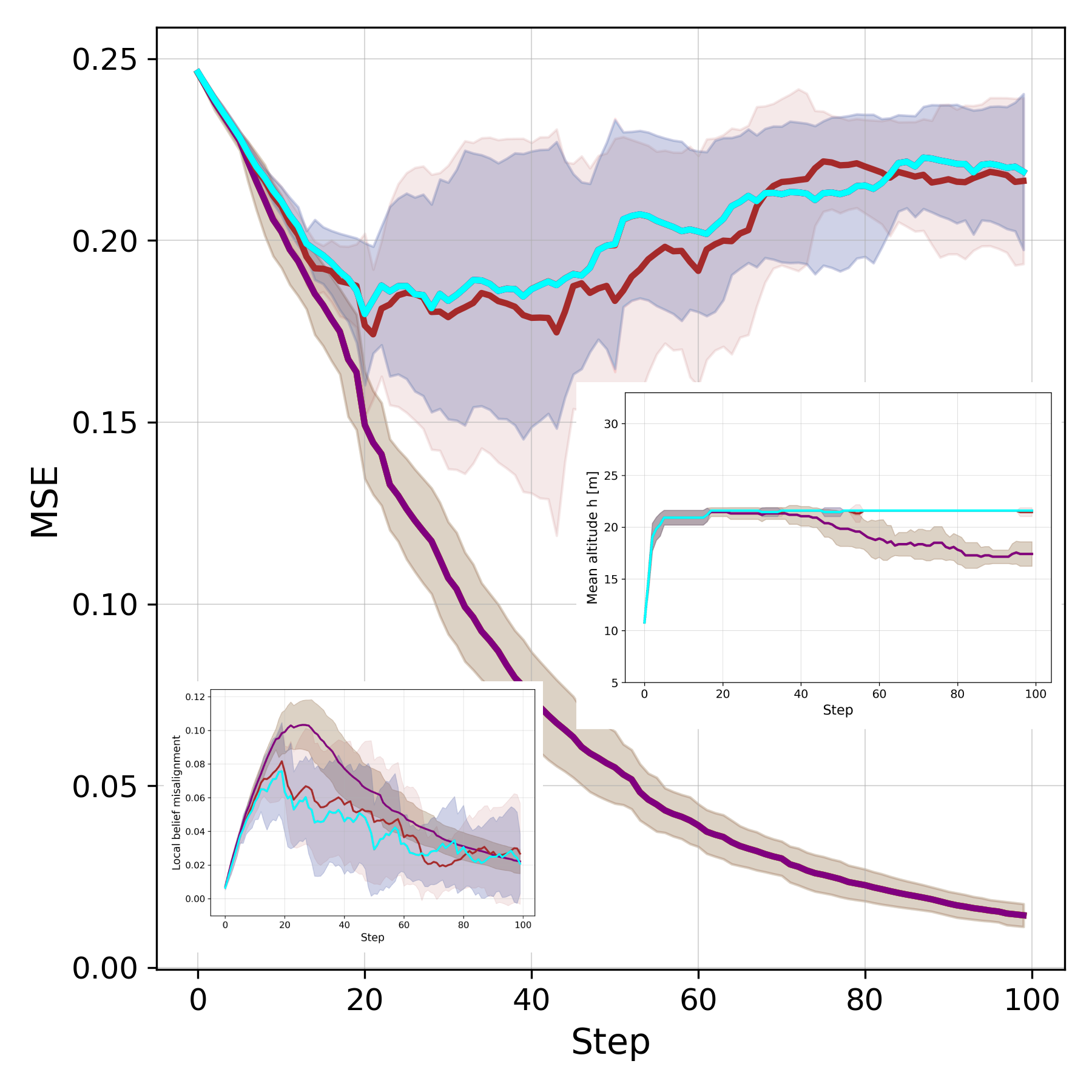}
        \label{fig:fig5_ig_bs_r5}
    }
    \hfill
    \subfloat[IGd, BS, R=5.]{
        \includegraphics[width=0.47\columnwidth]{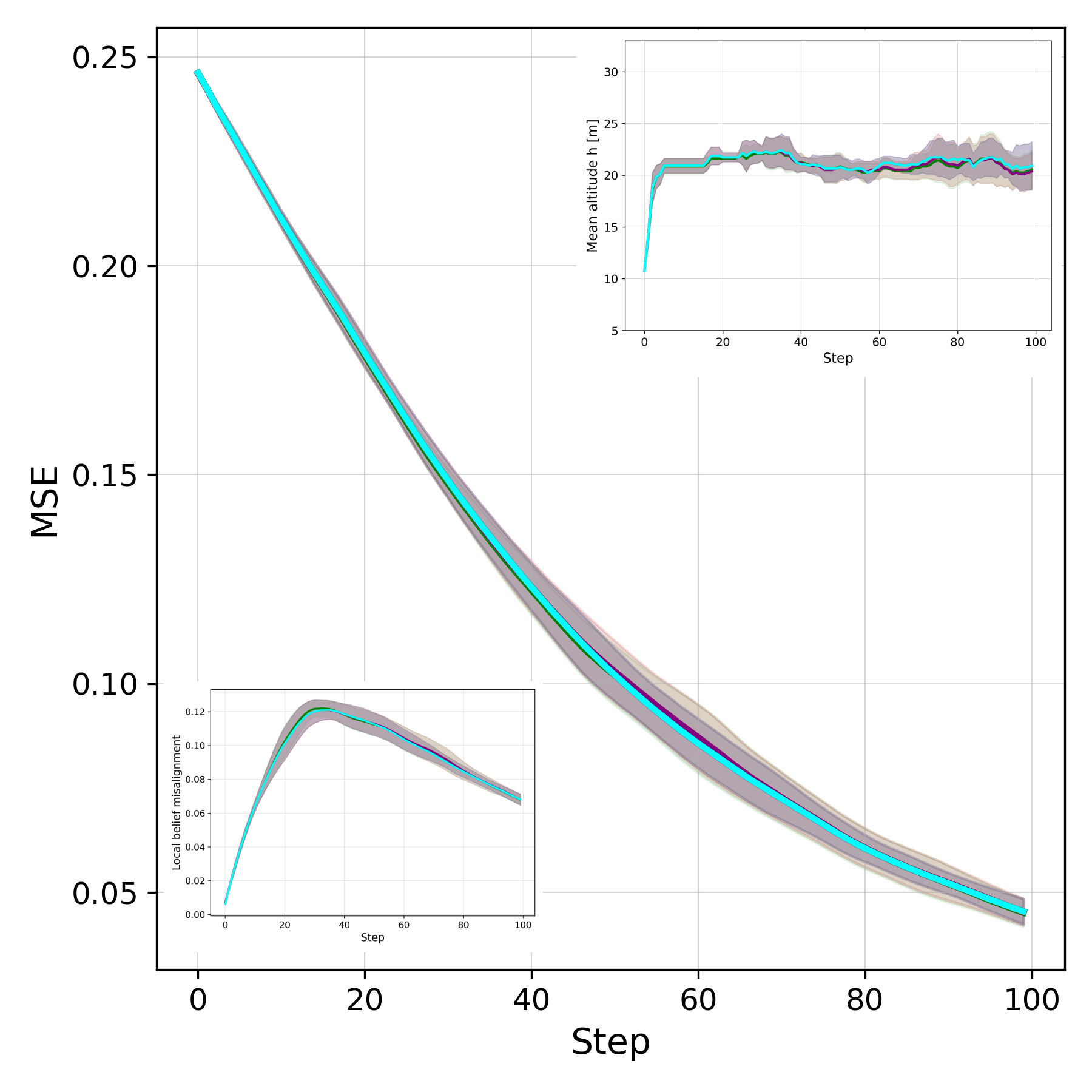}
        \label{fig:fig5_igd_bs_r5}
    }

    \vspace{0.08cm}

    \subfloat[IG, BM, R=5.]{
        \includegraphics[width=0.47\columnwidth]{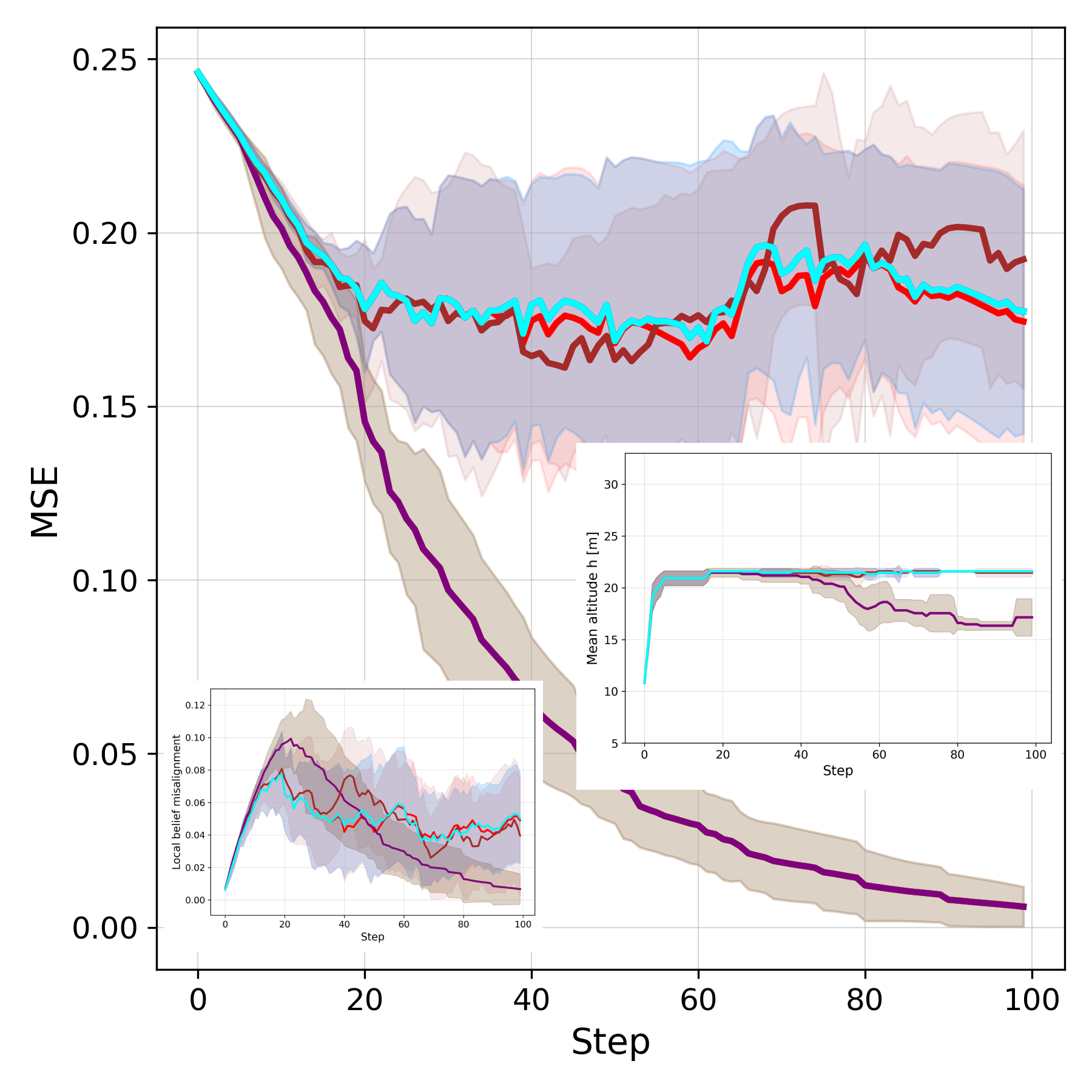}
        \label{fig:fig5_ig_bm_r5}
    }
    \hfill
    \subfloat[IGd, BM, R=5.]{
        \includegraphics[width=0.47\columnwidth]{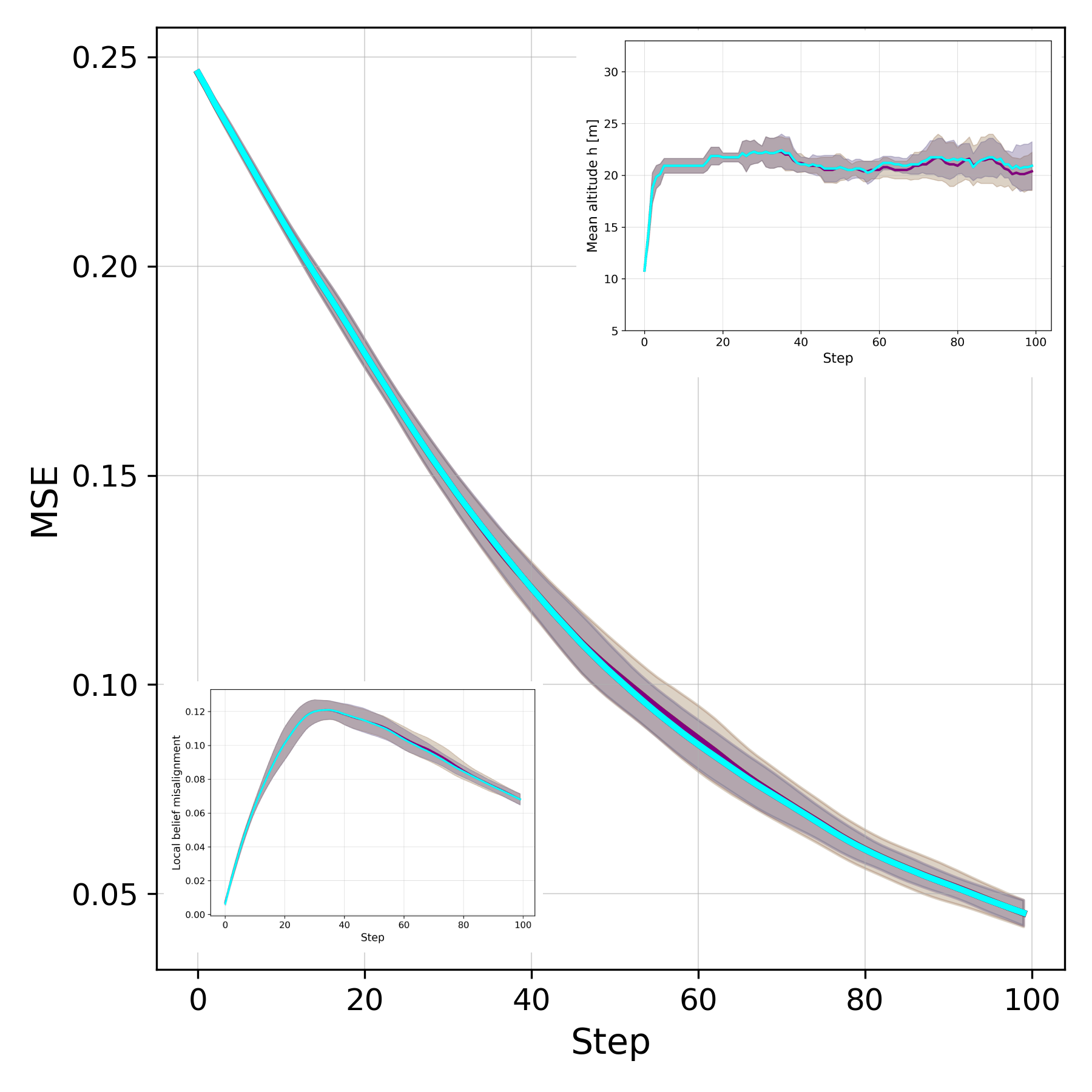}
        \label{fig:fig5_igd_bm_r5}
    }

    \caption{
    Comparison of different belief-fusion strategies for multi-UAV terrain mapping with \(N=4\) and communication range \(R=5\). The main plots show the MSE evolution, while the inset plots show the corresponding mean altitude and local belief misalignment.
    }
    \label{fig:fig5_fusion_methods_r5}
\end{figure}

\begin{figure}[t]
    \centering
    \subfloat[IG, BS, R=25.]{
        \includegraphics[width=0.47\columnwidth]{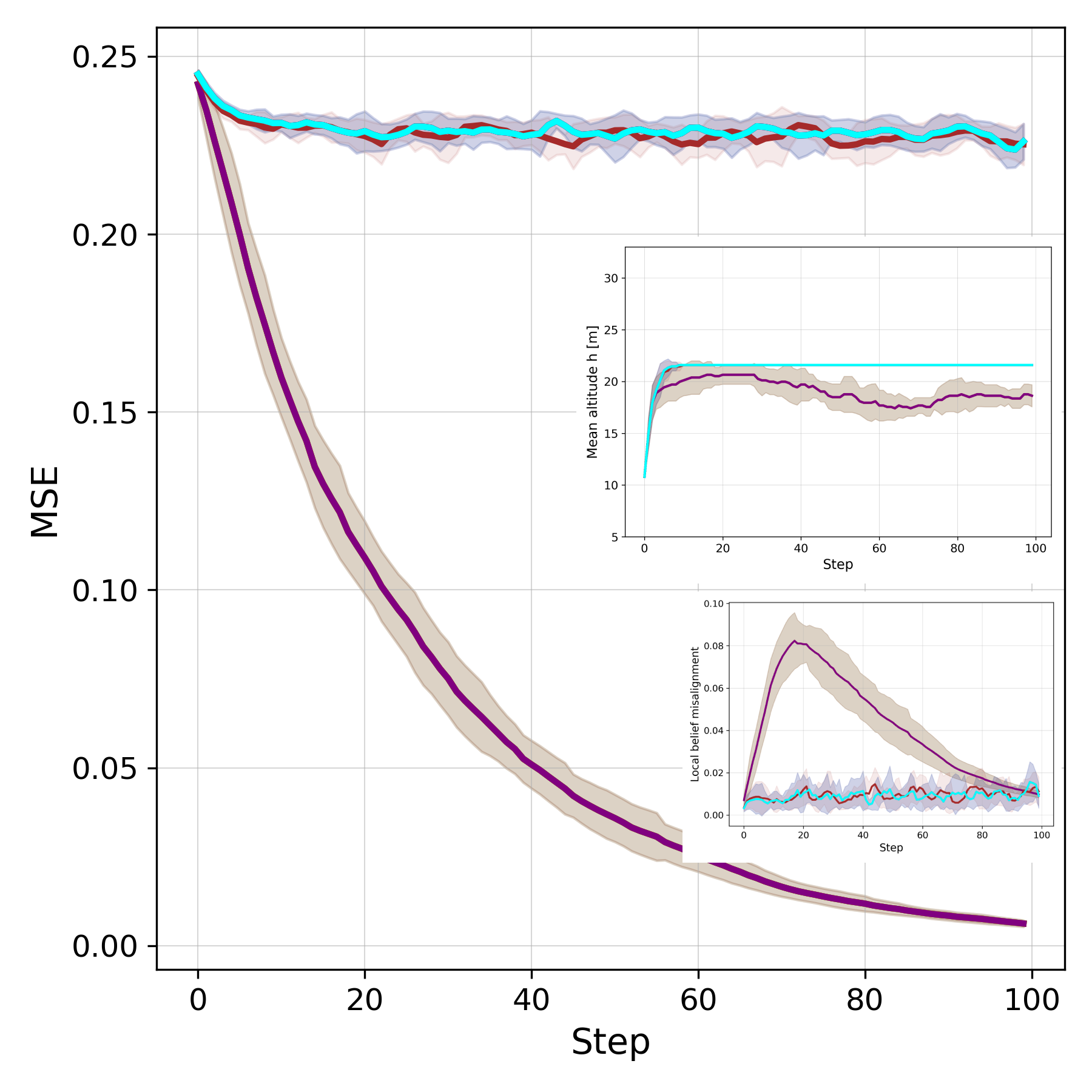}
        \label{fig:fig5_ig_bs_r25}
    }
    \hfill
    \subfloat[IGd, BS, R=25.]{
        \includegraphics[width=0.47\columnwidth]{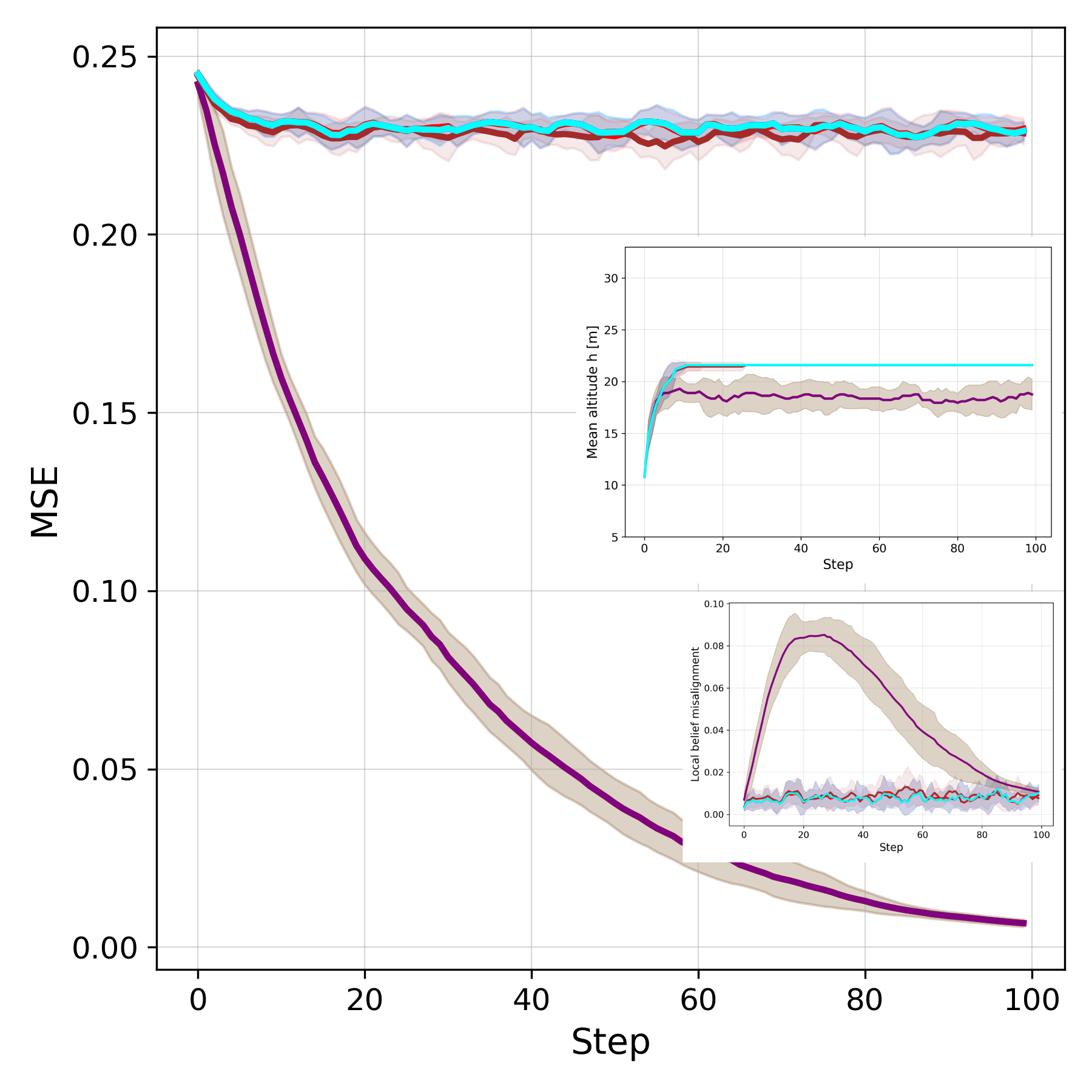}
        \label{fig:fig5_igd_bs_r25}
    }

    \vspace{0.08cm}

    \subfloat[IG, BM, R=25.]{
        \includegraphics[width=0.47\columnwidth]{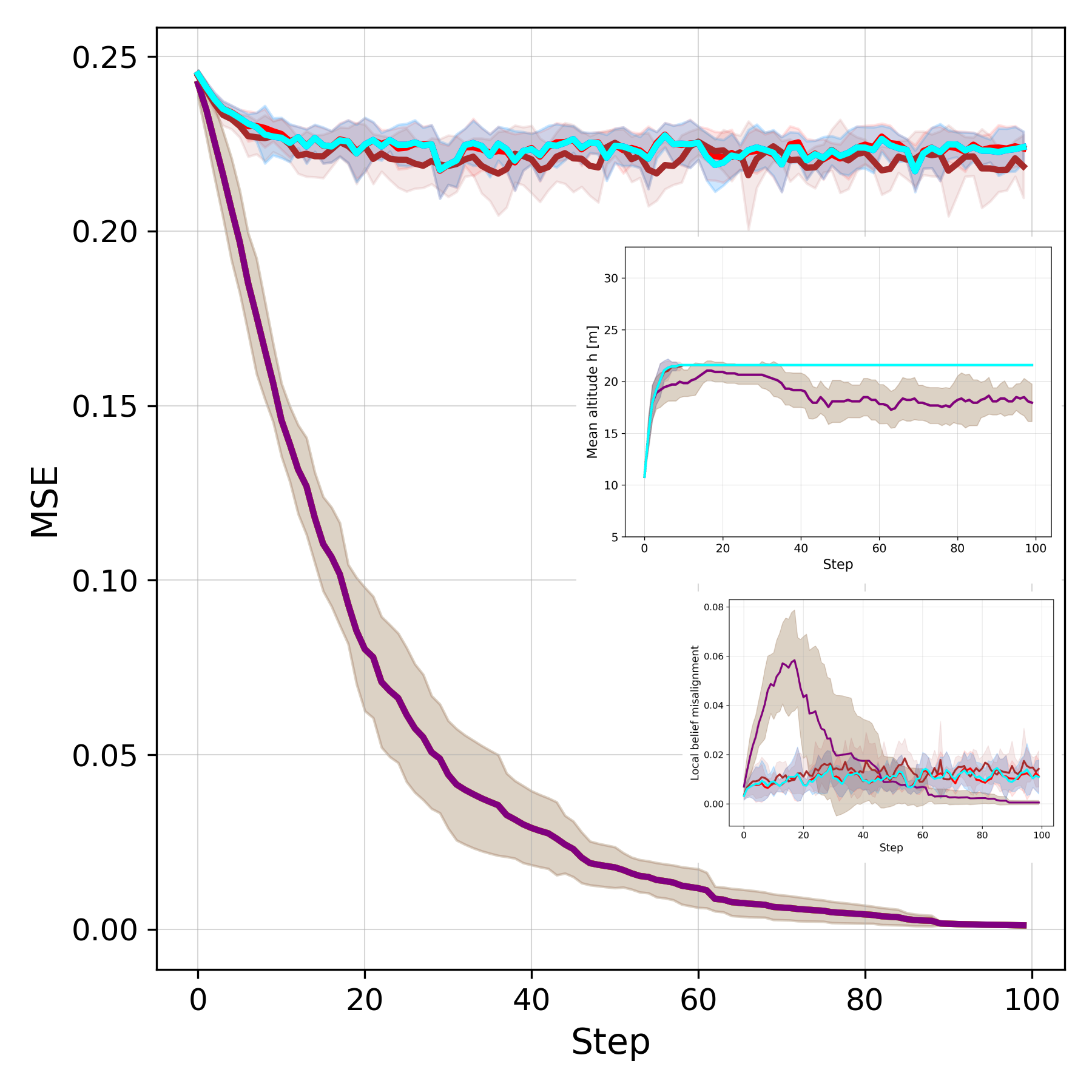}
        \label{fig:fig5_ig_bm_r25}
    }
    \hfill
    \subfloat[IGd, BM, R=25.]{
        \includegraphics[width=0.47\columnwidth]{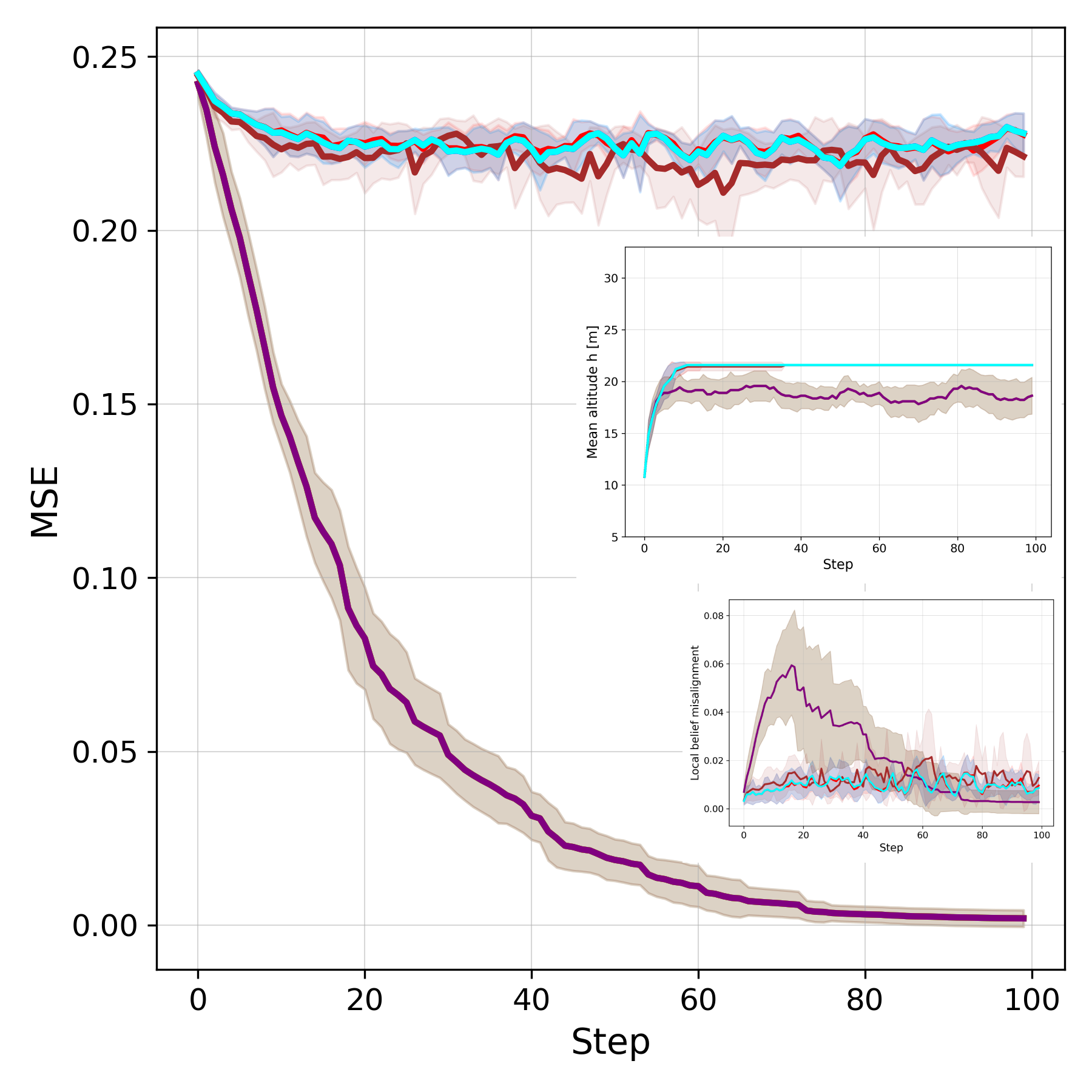}
        \label{fig:fig5_igd_bm_r25}
    }

    \caption{
    Comparison of different belief-fusion strategies for multi-UAV terrain mapping with \(N=4\) and communication range \(R=25\). The main plots show the MSE evolution, while the inset plots show the corresponding mean altitude and local belief misalignment.
    }
    \label{fig:fig5_fusion_methods_r25}
\end{figure}

\subsection{Effect of Information Sharing and Fusion}

This experiment evaluates the effect of information sharing and probabilistic fusion on real-world multi-UAV terrain mapping. A team of \(N=4\) UAVs is used, and two communication ranges are considered, namely \(R=5\) and \(R=25\). The IG and IGd planners are compared under three sharing modes: no sharing, single-news-belief sharing (BS), and multiple-news-belief sharing (BM). In BS, one recent news belief is shared and fused with neighboring UAVs, while in BM, separate news beliefs are maintained for different neighbors so that only the exchanged information is reset after communication. The evaluated fusion rules are Bayesian, weighted average, log-odds, confidence-weighted, Dempster--Shafer, covariance intersection, and consensus. The analysis focuses on MSE, local belief misalignment, and mean altitude.

\begin{table}[!t]
\centering
\caption{Representative final MSE values for information sharing and fusion.}
\label{tab:fusion_summary}
\scriptsize
\setlength{\tabcolsep}{3pt}
\begin{tabular}{llcccc}
\hline
Planner & Fusion group & BS \(R=5\) & BM \(R=5\) & BS \(R=25\) & BM \(R=25\) \\
\hline
 & Bayes/LO/DS     & 0.0144 & 0.0060 & 0.0063 & 0.0011 \\
IG  & Weighted/Cons.  & 0.2189 & 0.1774 & 0.2261 & 0.2241 \\
 & Cov. inter.     & 0.2164 & 0.1923 & 0.2253 & 0.2186 \\
\hline
 & Bayes/LO/DS     & 0.0451 & 0.0454 & 0.0068 & 0.0020 \\
IGd & Weighted/Cons.  & 0.0455 & 0.0455 & 0.2290 & 0.2279 \\
 & Cov. inter.     & 0.0455 & 0.0455 & 0.2284 & 0.2213 \\
\hline
\end{tabular}
\end{table}

Table~\ref{tab:fusion_summary} reports representative final MSE values for the main fusion-rule groups. Since Bayesian, log-odds, and Dempster--Shafer produced identical or nearly identical results in most cases, they are grouped as Bayes/LO/DS. Similarly, weighted-average and consensus are grouped because they showed very similar behaviour. The table shows that Bayes/LO/DS clearly provides the lowest final MSE, especially with BM and larger communication range. The other fusion groups do not improve the final MSE in the tested real-world experiments.

Fig.~\ref{fig:fig5_fusion_methods_none} shows the no-sharing case for IG and IGd. In this case, all fusion methods produce identical behaviour because no inter-UAV belief exchange is performed. Therefore, the no-sharing result is used as the baseline. For IG, the final MSE is \(0.0403\), the final local belief misalignment is \(0.0632\), and the final mean altitude is \(19.44\,\mathrm{m}\). For IGd, the final MSE is \(0.0454\), the final misalignment is \(0.0678\), and the final mean altitude is \(21.06\,\mathrm{m}\). This shows that, without belief sharing, the UAVs keep relatively independent local maps and the final misalignment remains high.

Fig.~\ref{fig:fig5_fusion_methods_r5} shows the results for the limited communication range \(R=5\). For the IG planner, information sharing clearly improves the mapping performance. With BS, Bayesian, log-odds, and Dempster--Shafer fusion reduce the final MSE to \(0.0144\) and the final misalignment to \(0.0221\). With BM, the performance improves further, reaching a final MSE of \(0.0060\) and a final misalignment of \(0.0067\). In contrast, weighted-average, consensus, confidence-weighted, and covariance-intersection fusion remain much less effective, with final MSE values around \(0.174\)--\(0.192\) under BM. This indicates that lower disagreement between UAVs does not always mean a more accurate map, because the agents may converge toward a consistent but inaccurate belief.

For the IGd planner with \(R=5\), the effect of sharing is much weaker. The final MSE remains approximately \(0.0451\)--\(0.0455\), and the final misalignment remains around \(0.068\) for both BS and BM. Therefore, under limited communication range, IGd does not substantially benefit from belief sharing. This suggests that the footprint-discounting mechanism in IGd may reduce redundant sensing, but it does not necessarily improve belief fusion when communication opportunities are limited.

Fig.~\ref{fig:fig5_fusion_methods_r25} shows the results for the larger communication range \(R=25\). Increasing the communication range significantly improves the best-performing fusion rules. For IG, Bayesian, log-odds, and Dempster--Shafer fusion achieve a final MSE of \(0.0063\) with BS and \(0.0011\) with BM. The corresponding final misalignment values are \(0.0097\) and \(0.0005\), respectively. For IGd, the same fusion family also becomes effective when \(R=25\), reducing the final MSE to \(0.0068\) with BS and \(0.0020\) with BM. The final misalignment values are \(0.0107\) and \(0.0027\), respectively. These results show that a larger communication range improves cooperative mapping and that BM generally provides better performance than BS.

The altitude results also show a consistent pattern. The best-performing fusion rules, namely Bayesian, log-odds, and Dempster--Shafer, usually lead to lower final mean altitudes, around \(17\)--\(19\,\mathrm{m}\). In contrast, the weaker fusion rules often keep the UAVs near the maximum sensing altitude of approximately \(21.6\,\mathrm{m}\). This suggests that when shared beliefs become more informative, the UAVs can descend for local refinement. When fusion is ineffective, the UAVs remain in a high-altitude exploration mode and the MSE reduction is limited.

Overall, the results show that information sharing and fusion strongly affect multi-UAV mapping performance. Bayesian, log-odds, and Dempster--Shafer fusion consistently provide the best results and show almost identical behaviour in this experiment. BM outperforms BS, especially when the communication range is large. In contrast, weighted-average, consensus, confidence-weighted, and covariance-intersection fusion do not provide reliable improvement in the tested real-world setting. The results also show that local belief misalignment should be interpreted together with MSE, because low misalignment alone does not guarantee high mapping accuracy.

\section{Conclusion}

This paper highlighted a real-world validation of a multi-UAV active sensing framework for precision agriculture. The proposed framework combines probabilistic belief mapping, altitude-dependent sensing, information-gain-based planning, spatial correlation modeling, and multi-UAV information sharing for binary vegetation/no-vegetation terrain mapping. The evaluation was performed using both synthetic correlated terrains and real UAV-derived agricultural imagery.

The results demonstrate that the Information Gain (IG) planner provides the best overall performance compared with Random Walk and Sweep baselines. By selecting actions according to the current uncertainty of the belief map, IG reduces entropy and MSE more effectively in both synthetic and real-world experiments. Sweep remains a strong baseline because it provides systematic field coverage, but it cannot adapt to spatial uncertainty. Random Walk performs weakest because its motion is not guided by either coverage or information value. The real-world experiments also show that sensing geometry is critical: increasing the FoV from \(30^\circ\) to \(46^\circ\) improves map accuracy by increasing the observable footprint while preserving the adaptive behavior of the IG planner. The spatial correlation analysis shows that pairwise modeling influences both uncertainty reduction and mapping accuracy. In the tested real-world terrain, equal weights provide the strongest entropy reduction, while biased weights achieve the lowest final MSE. The adaptive sigmoid and hyperbolic tangent weights show similar behavior and do not improve over the simpler strategies, suggesting that adaptive correlation estimation can be sensitive to irregular vegetation boundaries, shadows, and noisy local observations.

The multi-UAV experiments further confirm that information sharing and probabilistic fusion are important for cooperative mapping. Bayesian, log-odds, and Dempster--Shafer fusion achieve the best results and behave almost identically under the adopted binary belief representation. Belief-map sharing generally outperforms belief-state sharing, and increasing the communication range from \(R=5\) to \(R=25\) improves both final MSE and local belief alignment. However, low belief misalignment alone does not guarantee high map accuracy, since some fusion rules can lead to consistent but inaccurate beliefs. The proposed framework is also related to broader concepts in information-driven search and cooperative sensing. The IG-based planner follows the same principle as infotaxis by selecting actions that maximize expected information acquisition. The synthetic terrains highlight the role of spatial structure, where increasing the cluster radius reduces spatial disorder and increases spatial correlation. The comparison between IG, Sweep, and Random Walk is consistent with the No Free Lunch principle: no planner is universally optimal, and performance depends on the sensing objective, terrain structure, and sensing constraints. Finally, the multi-UAV experiments introduce a cooperative, swarm-like sensing setting, where distributed belief sharing improves mapping performance without requiring a fully centralized controller. Ultimately, the study shows that effective real-world UAV-based agricultural mapping requires the joint design of decision making, sensing altitude, spatial modeling, and information sharing.

Future work will investigate cooperative centralized decision making, and learning-based planning strategies to improve scalability, communication efficiency, and robustness in both simulation and real-world multi-UAV missions.

\end{document}